\pdfoutput=1

\documentclass[11pt]{article}

\usepackage[final]{acl}

\usepackage{times}
\usepackage{latexsym}

\usepackage[T1]{fontenc}

\usepackage[utf8]{inputenc}

\usepackage{microtype}

\usepackage{inconsolata}

\usepackage{graphicx}

\usepackage{amsmath}
\usepackage{amssymb}
\usepackage{amsfonts}
\usepackage{multirow}
\usepackage{booktabs}
\usepackage{adjustbox}

\title{Unveiling Language-Specific Features in Large Language Models \\via Sparse Autoencoders}

\author{Boyi Deng\textsuperscript{1},
	Yu Wan\thanks{Corresponding authors}, 
        Yidan Zhang\textsuperscript{2},
        Baosong Yang,
        Fuli Feng\textsuperscript{1}$^*$
 \\
    \textsuperscript{1}University of Science and Technology of China,\\
    \textsuperscript{2}Sichuan University\\
	\texttt{dengboyi@mail.ustc.edu.cn}\\
 }

\begin{document}
\maketitle
\begin{abstract}
The mechanisms behind multilingual capabilities in Large Language Models (LLMs) have been examined using neuron-based or internal-activation-based methods. However, these methods often face challenges such as superposition and layer-wise activation variance, which limit their reliability. Sparse Autoencoders (SAEs) offer a more nuanced analysis by decomposing the activations of LLMs into a sparse linear combination of SAE features. We introduce a novel metric to assess the monolinguality of features obtained from SAEs, discovering that some features are strongly related to specific languages. Additionally, we show that ablating these SAE features only significantly reduces abilities in one language of LLMs, leaving others almost unaffected. Interestingly, we find some languages have multiple synergistic SAE features, and ablating them together yields greater improvement than ablating individually. Moreover, we leverage these SAE-derived language-specific features to enhance steering vectors, achieving control over the language generated by LLMs. The code is publicly available at~\url{https://github.com/Aatrox103/multilingual-llm-features}.

\end{abstract}

\section{Introduction}
Large Language Models (LLMs)~\citep{openai2024gpt4technicalreport,grattafiori2024llama3herdmodels,qwen2025qwen25technicalreport} exhibit impressive abilities in various domains such as text generation~\citep{openai2024gpt4technicalreport,grattafiori2024llama3herdmodels,xu2025personalized}, instruction following~\citep{zhang2024instructiontuninglargelanguage,lou2024largelanguagemodelinstruction}, and reasoning~\citep{DBLP:conf/acl/0009C23}. Recently, considerable efforts have been made to enhance the multilingual capabilities of LLMs to meet the growing demand for their deployment in multilingual environments~\citep{qin2024multilinguallargelanguagemodel,huang2025surveylargelanguagemodels}. For instance, Gemini 1.5 incorporates a variety of multilingual data in its training process and emphasizes its multilingual capabilities~\citep{geminiteam2024gemini15unlockingmultimodal}.~\citet{yang2024qwen2technicalreport} claim that Qwen2 supports over 30 languages and achieves great performance on multilingual benchmarks. Moreover, multilingual training data comprises approximately 3\% of the training data for Llama 3, and there are also high-quality multilingual instruction-tuning data for 8 languages~\citep{grattafiori2024llama3herdmodels}. As the significance of multilingual capabilities in LLMs continues to grow, it is crucial to delve into the mechanisms of these capabilities to enhance them further.

Works focusing on the mechanisms of multilingual capabilities in LLMs can be broadly divided into neuron-based and internal-activation-based methods. Neuron-based methods aim to identify language-specific neurons and analyze their impact on the corresponding language~\citep{DBLP:conf/acl/ZhangZ0G024,DBLP:conf/nips/0006ZCKB24,DBLP:conf/acl/TangLH0WZWW24,DBLP:conf/naacl/KojimaOIYM24}. And activation-based method attempts to obtain token distributions at intermediate layers using the unembedding matrix in the final layer~\citep{zhong2024englishcentricllmslanguagemultilingual,DBLP:conf/acl/WendlerVM024}. 
However, neuron-based methods are sometimes unreliable, due to ``superposition''~\citep{elhage2022toymodelssuperposition}, which suggests that neural networks often consolidate multiple unrelated concepts into a single neuron.
Additionally, activation-based method often has significant errors except in the last few layers, due to the varying distribution of activations across different layers. As such, it is important to use a more reliable and interpretable method to analyze multilingual capabilities in LLMs.

To achieve this, we use Sparse Autoencoders (SAEs)~\citep{bricken2023monosemanticity,cunningham2023sparse}, which are designed to decompose language model activations in each layer into a sparse linear combination of SAE features. The advantages of SAEs in analyzing multilingual capabilities in LLMs are threefold. First, SAEs can be applied to individual tokens, providing a more monosemous analysis compared to neuron-based methods. Second, SAEs are trained on each layer separately, making them more reliable when analyzing activations from different layers than current activation-based methods. Third, multilingual data is naturally parallel, meaning that ideally, the main difference between multilingual data is the language, so it is easy to identify monolingual features with SAEs.

Given the advantages of SAEs, we use them to analyze multilingual capabilities in LLMs. Concretely, we start with a preliminary experiment in which we find high activation of some features in a certain language. Inspired by this, we propose a metric to measure the monolinguality of a feature based on the activation difference across different languages. The results show that some features possess strong monolingual characteristics. Moreover, we believe that these language-specific features are not only related to language-specific tokens, so we experiment on a ``code-switching''~\citep{kuwanto2024linguistics,winata-etal-2023-decades} dataset and find that language-specific features are also closely associated with the language-specific linguistic context. Furthermore, we use \textit{directional ablation}~\citep{NEURIPS2024_f5454485,ferrando2024iknowentityknowledge} to ``zero out'' language-specific features during the forward pass of LLMs, resulting in a loss of capabilities in only certain language. Interestingly, we observe that some languages may exhibit more than one specific feature. And these features have a synergistic relationship, meaning ablating these features together results in a significant improvement compared to ablating them individually.

The language-specific features we find are of great monolinguality, so we further leverage them to improve \textit{steering vectors}~\citep{DBLP:journals/corr/abs-2308-10248}. Concretely, we use language-specific features as gating signals to control steering vectors and achieve better control over the language generated by LLMs, which validates the practical potential of these language-specific features.

In summary, our main contributions are:
\begin{itemize}
\item We use SAE, a more human-interpretable method, to analyze multilingual capabilities of LLMs, and propose a metric to measure the monolinguality of SAE features.
\item We find some SAE features that are not only related to language-specific tokens but also related to language-specific linguistic context.
\item We find that ablating language-specific features only significantly decreases the language-specific capabilities of LLMs. 
\item We use language-specific features as gating signals to improve \textit{steering vectors}, and achieve better control over the language generated by LLMs.
\end{itemize}

\section{Preliminary}
\paragraph{SAEs.}

\begin{figure*}[t] 
\setlength{\abovecaptionskip}{-0.10cm}
\setlength{\belowcaptionskip}{-0.30cm}
\centering
  \includegraphics[width=\textwidth]{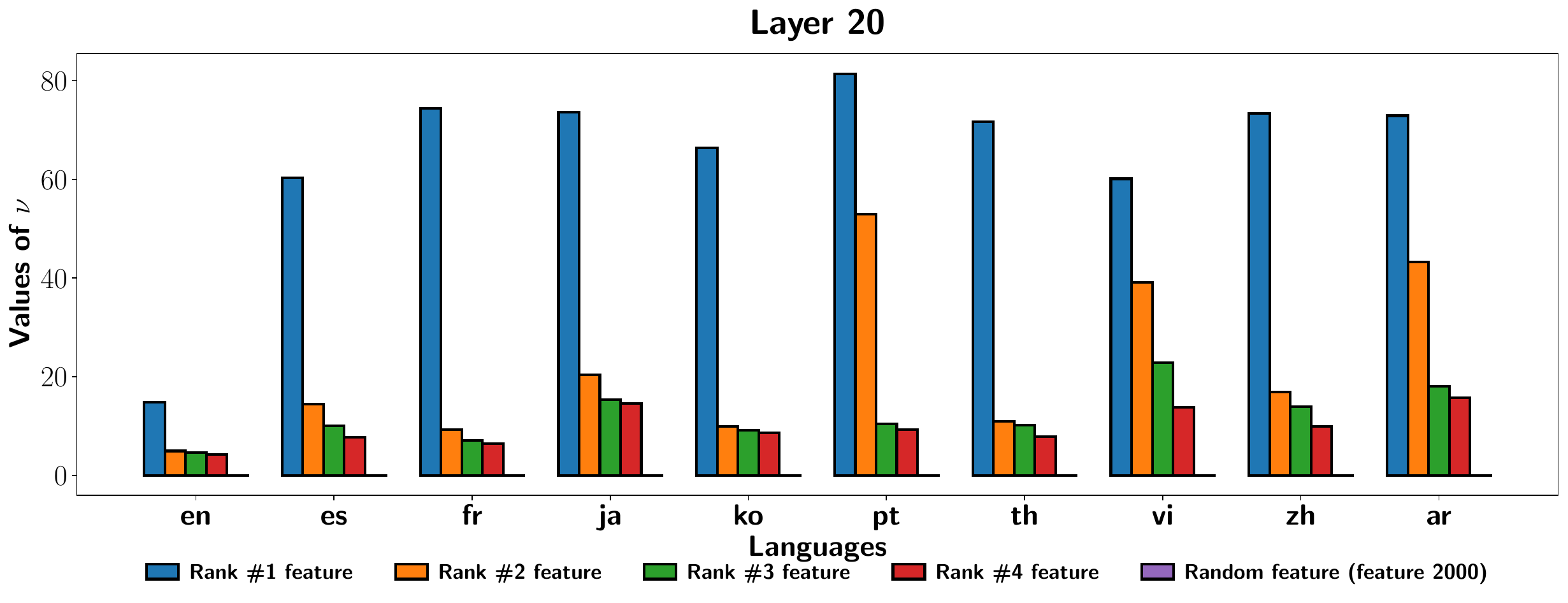}
  \caption{The values of $\nu$, as referenced in Eq.~\ref{eq:ablation}, where a larger $\nu$ indicates stronger monolingualism, are reported for the top-4 features and a random feature across various languages in layer 20 of Gemma 2 2B. The values of $\nu$ for the top-4 features are greater than those of a random feature. In most languages, the top-1 feature possesses a significantly larger $\nu$. Additional results for other layers and LLMs are in Appendix~\ref{Appendix:nu}, exhibiting similar patterns. The value of the random feature (feature 2000) is too small to be visible.
} 
  \label{fig:histogram_gemma-2-2b_layer_20}
\end{figure*}

SAEs are a specialized form of autoencoders~\citep{autoencoder1} designed to decompose language model activations into a sparse linear combination of learned feature directions. Given a language model activation $\mathbf{x} \in \mathbb{R}^N$ in certain layer\footnote{We use the residual stream at each layer as $\mathbf{x}$ because it is more interpretable~\citep{ferrando2024iknowentityknowledge,chanin2024absorption}.}, the SAE computes a feature activation $\mathbf{f} \in \mathbb{R}^M$, where $M \gg N$, and reconstructs the input as $ \hat{\mathbf{x}} $. The typical reconstruction process is described by the equations:
\begin{align}
    \mathbf{f(x)} &:=  \text{ReLU}(\mathbf{W_{\text{enc}}}\mathbf{x} + \mathbf{b_{\text{enc}}}), \label{eq:relu}\\
    \hat{\mathbf{x}}(\mathbf{f}) &:= \mathbf{W_{\text{dec}}}\mathbf{f} + \mathbf{b_{\text{dec}}}. \label{eq:reconstruct}
\end{align}
To ensure that $\mathbf{f}$ remains sparse,~\citet{bricken2023monosemanticity,cunningham2023sparse} incorporate an L1 penalty on $\mathbf{f}$ into the training loss function. Another approach by~\citet{gao2024scaling} employs Top-K SAEs, which enforce sparsity by selecting only the K most active dimensions of $\mathbf{f}$, setting all the others to zero. Following the notation of \citet{rajamanoharan2024jumpingaheadimprovingreconstruction}, we denote the columns of $\mathbf{W_{\text{dec}}}$ as $\mathbf{d}_i$ for $i=1, \ldots, M$. These columns represent the feature directions into which the SAE decomposes the vector $\mathbf{x}$. For simplicity, we will refer to each column as a ``feature'' throughout this paper.

\paragraph{Datasets.}
Flores-200~\citep{flores1,flores2} is a parallel corpus that contains translations of English sentences into 200 different languages. 
Due to the semantic similarity of the translated sentences, this dataset is particularly useful for comparing linguistic features across languages. 
We extract a subset called Flores-10, which includes 10 languages\footnote{English (en), Spanish (es), French (fr), Japanese (ja), Korean (ko), Portuguese (pt), Thai (th), Vietnamese (vi), Chinese (zh), and Arabic (ar).}.

\begin{figure}[t] 
\setlength{\abovecaptionskip}{-0.05cm}
\setlength{\belowcaptionskip}{-0.30cm}
\centering
  \includegraphics[width=\columnwidth]{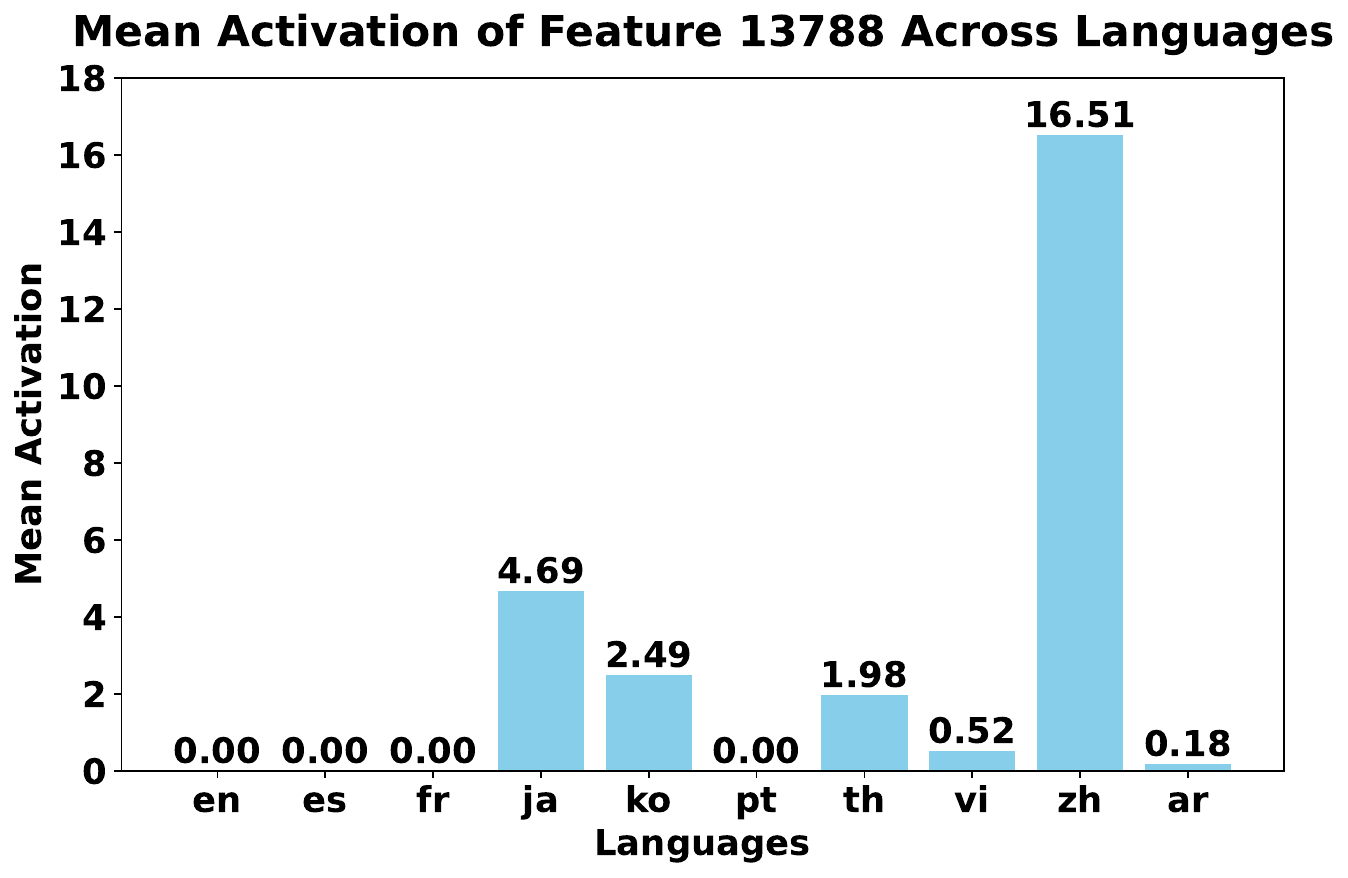}
  \caption{The mean activation of feature 13788 across different languages in layer 10 of Gemma 2 2B. The high mean activation in Chinese suggests that feature 13788 might be related to Chinese.} 
  \label{fig:activation_per_lan}
\end{figure}
\paragraph{Models.}
To ensure the robustness of our findings, we include a diverse set of LLMs and their corresponding SAEs. We use SAEs from Gemma Scope~\citep{lieberum2024gemmas} for Gemma 2 2B and Gemma 2 9B~\citep{team2024gemma}, and SAEs from Llama Scope~\citep{he2024llamas} for Llama-3.1-8B. 


\section{Language-Specific Features}\label{sec:lan_specific_feature}

\subsection{Finding Language-Specific Features}
To find language-specific features, we conduct a preliminary experiment by prompting Flores-10 into the LLMs and analyzing the residual stream using SAEs. We find that the mean activation of some features is particularly high for a certain language, while remaining very low for other languages, as illustrated by the example in Figure~\ref{fig:activation_per_lan}. Inspired by this, we propose a metric to measure the monolinguality of a feature. Specifically, given a set $\mathcal{D} = \{\mathcal{D}_1, \ldots, \mathcal{D}_K\}$, which contains the residual stream set for a certain layer with $K$ different languages, we calculate the mean activation difference of feature $s$ for a specific language $L$ compared to the other languages as follows:
\begin{align}
    \mu^L_s &= \frac{1}{|\mathcal{D}_L|}\sum_{\mathbf{x}\in\mathcal{D}_L}\mathbf{f}_s(\mathbf{x}),\notag\\
    \gamma^L_s &= \frac{1}{|\mathcal{D}\setminus\{\mathcal{D}_L\}|}\sum_{\mathcal{D}_I\in \mathcal{D}\setminus\{\mathcal{D}_L\}}\frac{1}{|\mathcal{D}_I|}\sum_{\mathbf{x}\in\mathcal{D}_I}\mathbf{f}_s(\mathbf{x}),\notag\\
    \nu^L_s &= \mu^L_s-\gamma^L_s,\label{eq:AD} 
\end{align}
where $\mathbf{f}_s(\mathbf{x})$ is the activation of feature $s$. We calculate $\nu$ for all languages and features and rank them from high to low for each language. The top-ranked features are considered language-specific features.

\subsection{Monolinguality Analysis}
We use the first 100 data points in Flores-10 to calculate $\nu$ for each language. The results are shown in Figure~\ref{fig:histogram_gemma-2-2b_layer_20}. From this figure, we make the following observations. 
(1) The mean activation of the top-4 features is significantly higher than that of a random feature, which remains close to zero.
(2) For most languages, the mean activation of the top features decreases rapidly among the first few, and the mean activation of the rank \#1 feature is considerably higher than the others.
(3) In some languages, the rank \#2 feature also shows a substantially large mean activation compared to other features.
These results suggest that top-ranked features possess strong monolingual characteristics, and in most scenarios, the top-1 feature suffices in capturing these characteristics.\footnote{English is the primary language for most LLMs, and it often exhibits different characteristics compared to other languages~\citep{qin2024multilinguallargelanguagemodel}, so we only focus on non-English results in the subsequent sections.}

\begin{figure}[t] 
\setlength{\abovecaptionskip}{-0.05cm}
\setlength{\belowcaptionskip}{-0.30cm}
\centering
  \includegraphics[width=\columnwidth]{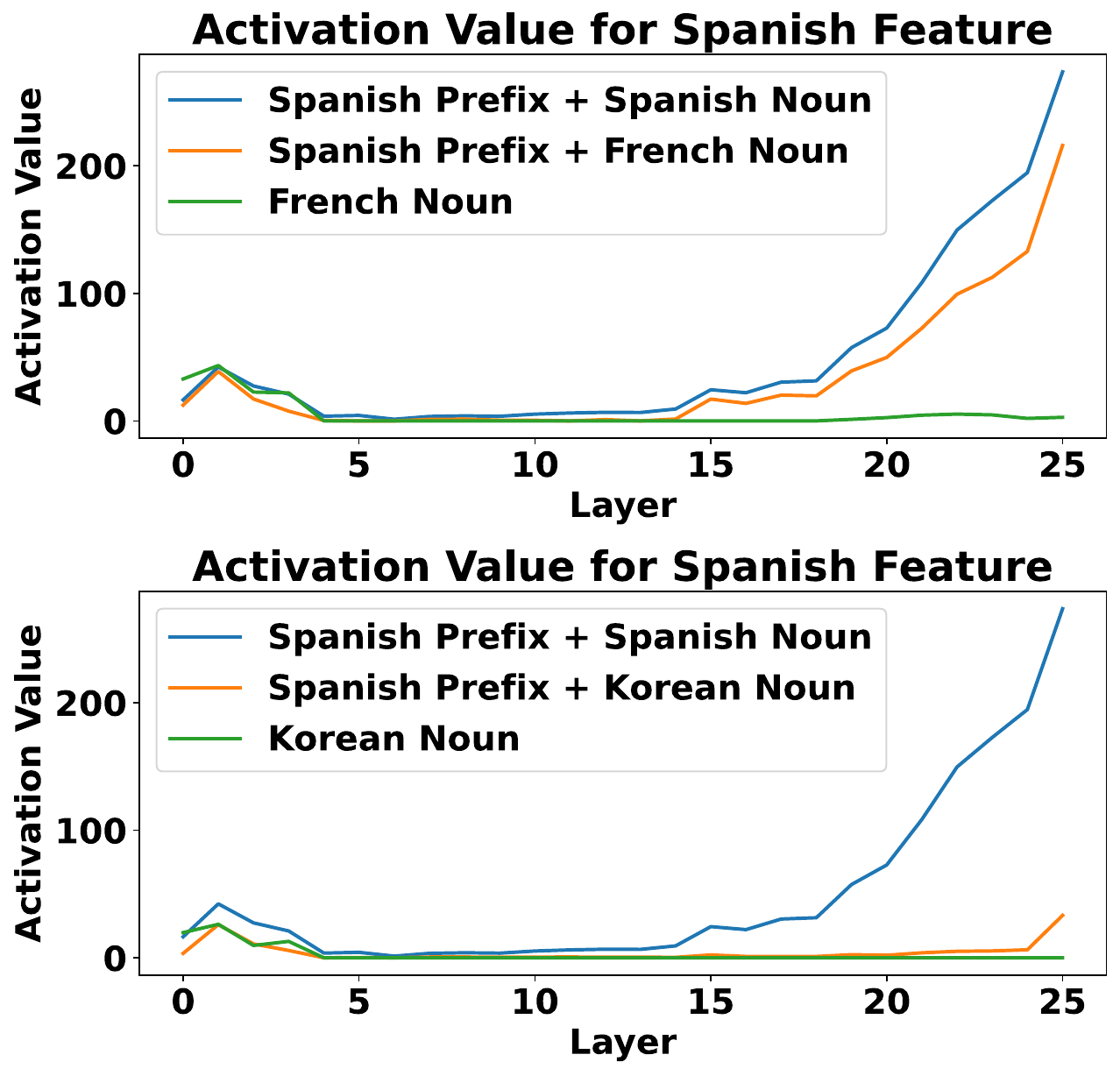}
  \caption{The mean activation values for the Spanish feature with various noun and prefix combinations. Adding a Spanish prefix enhances the Spanish feature activation for non-Spanish nouns, enabling the LLM to process them as if they were ``Spanish tokens.''} 
  \label{fig:codeswitch}
\end{figure}

\begin{figure}[t] 
\setlength{\abovecaptionskip}{-0.05cm}
\setlength{\belowcaptionskip}{-0.30cm}
\centering
  \includegraphics[width=\columnwidth]{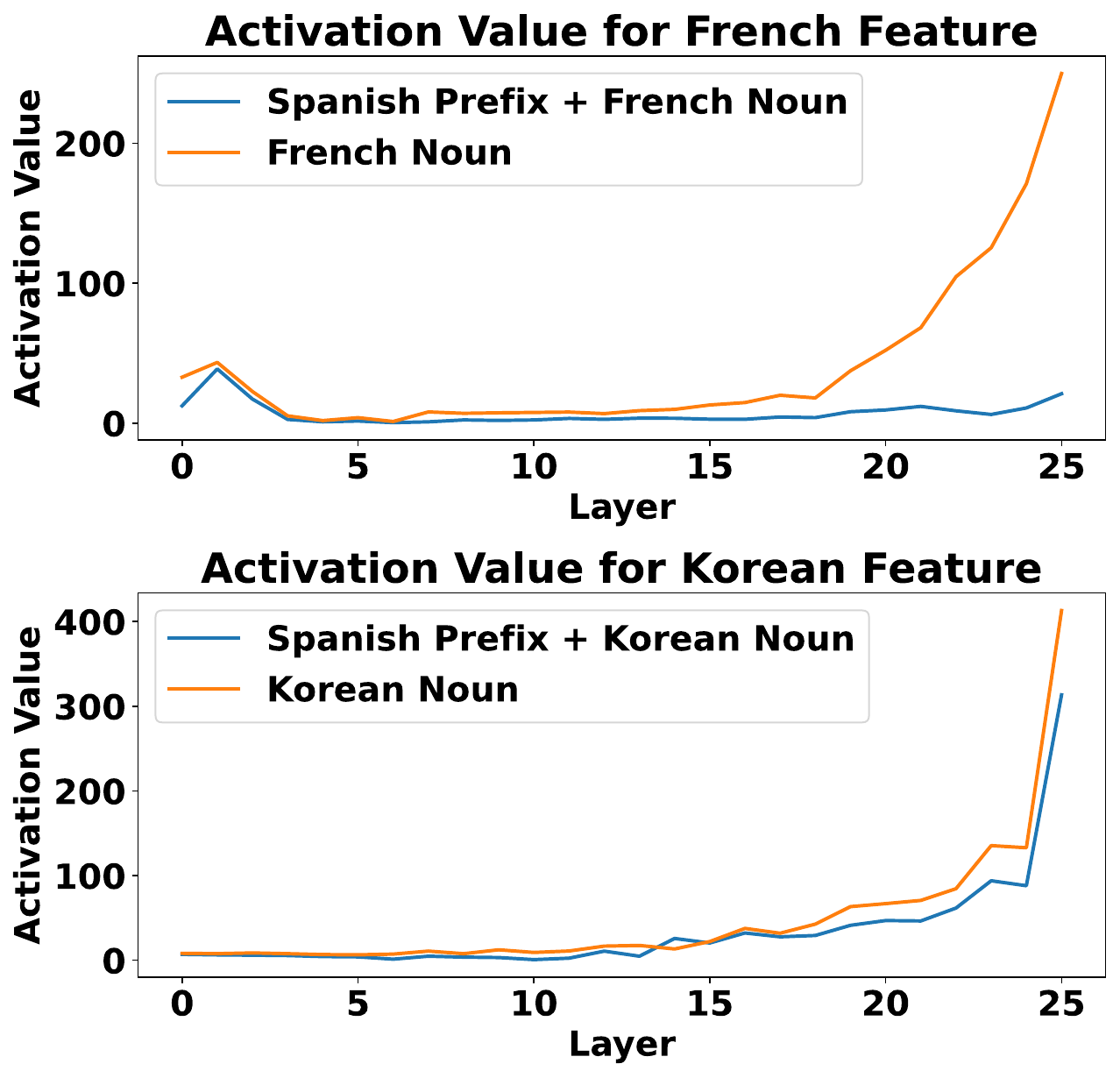}
  \caption{The mean activation values for the French and Korean features with various noun and prefix combinations. Introducing a different language prefix decreases the original language feature activation of nouns.} 
  \label{fig:codeswitch2}
\end{figure}

\begin{figure*}[t] 
\setlength{\abovecaptionskip}{-0.10cm}
\setlength{\belowcaptionskip}{-0.30cm}
\centering
  \includegraphics[width=\textwidth]{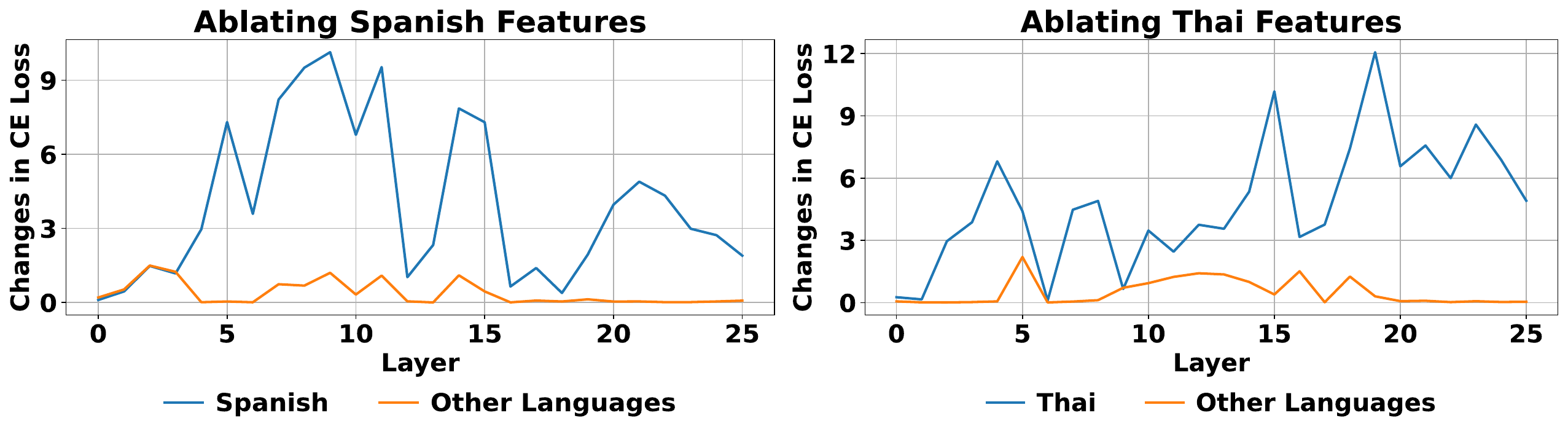}
  \caption{The changes in CE loss on texts in the target language and texts in other languages after ablating language-specific features. Ablating language-specific features has a much larger impact on the CE loss of texts in the target language compared to texts in other languages. We provide results for Gemma 2 2B here, additional results can be found in Appendix~\ref{Appendix:Directional_Ablation}.}
  \label{fig:celoss}
\end{figure*}

\section{Language-Specific Features Extend Beyond Language-Specific Tokens}
In earlier sections, we only evaluate language-specific features on monolingual texts. This raises a natural question: are these language-specific features solely related to language-specific tokens? To explore this, we focus on a phenomenon called ``code-switching.''\footnote{Code-switching refers to the practice of alternating between two or more languages within a single text~\citep{kuwanto2024linguistics,winata-etal-2023-decades}.} Our findings indicate that language-specific features are also related to language-specific linguistic context.

\subsection{Experimental Settings}
\paragraph{Code-Switching Dataset.} We use GPT-4o to generate sentences in various languages, each ending with a noun. We then replace the noun with its equivalent in other languages. For each language, we generate 5 simple sentences, and each sentence has 8 variants where the noun is substituted with its equivalent in different languages. Example data are shown in Figure~\ref{fig:cs_data}. We only report results of Gemma 2 2B for Spanish prefix, additional results with the same patterns are in Appendix~\ref{sec:cs_appendix}.
\paragraph{Metric}
To analyze the impact of different language prefixes on ending nouns, we calculate the mean activation of language-specific features for the ending nouns both with and without a prefix.

\subsection{Results}
\paragraph{Spanish Prefix Enhances Spanish Features in Non-Spanish Nouns.}
We analyze the mean activation values of the Spanish features for Spanish, French, and Korean nouns, comparing scenarios with and without Spanish prefixes, as illustrated in Figure~\ref{fig:codeswitch}. Our observations are as follows:
(1) Introduction of a Spanish prefix to a French or Korean noun results in higher Spanish feature activation values compared to when the French or Korean nouns stand alone. However, the value is still lower than that of the combination of Spanish prefixes and Spanish nouns.
(2) The activation value for Spanish features of stand-alone French and Korean nouns remains relatively low across all layers.
(3) Both French and Korean nouns with a Spanish prefix show greater increases in Spanish feature activations at deeper layers than at shallower ones.
(4) Adding a Spanish prefix results in a larger increase in the Spanish feature for French nouns compared to Korean nouns. 
These findings suggest that adding a Spanish prefix enhances the Spanish feature activation for non-Spanish nouns, enabling the LLM to process them as if they were ``Spanish tokens.'' Consequently, this allows the LLM to use these non-Spanish tokens within a consistent language context.

\paragraph{Spanish Prefix Decreases Non-Spanish Features in Non-Spanish Nouns.}

We also analyze the mean activation values of the French and Korean features for corresponding nouns, comparing scenarios with and without Spanish prefixes, as presented in the provided Figure~\ref{fig:codeswitch2}. Our observations are as follows:
(1) For French and Korean nouns, the original language feature activation is significantly higher when the nouns are standalone than when preceded by a Spanish prefix.
(2) Both French and Korean nouns show greater decreases in their original language feature activations at deeper layers than at shallower ones.
(3) Adding a Spanish prefix results in a larger decrease in the corresponding feature for French nouns compared to Korean nouns. 
These findings reveal that introducing a different language prefix decreases the original language feature activation of nouns, making them less like nouns from their original language.

\vspace{-2pt}
\paragraph{Language-Specific Features Extend Beyond Language-Specific Tokens.}

The results in Figures~\ref{fig:codeswitch} and \ref{fig:codeswitch2} suggest that language-specific features are not solely tied to specific language tokens but are also closely associated with the language-specific linguistic context. This suggests that the linguistic characteristics recognized by the model extend beyond individual words to encompass the contextual environment in which these words appear. Notably, the influence of a Spanish prefix is more pronounced on French nouns than on Korean nouns, potentially due to the linguistic similarities between Spanish and French. This highlights the model's ability to dynamically adapt its feature activations based on the surrounding linguistic context, effectively reinterpreting non-Spanish tokens within a Spanish framework while diminishing their original language attributes.

\section{Ablating Language-Specific Features Leads to Language-Specific Changes}
\label{sec:ablate}

\begin{figure*}[t] 
\setlength{\abovecaptionskip}{-0.10cm}
\setlength{\belowcaptionskip}{-0.50cm}
\centering
  \includegraphics[width=\textwidth]{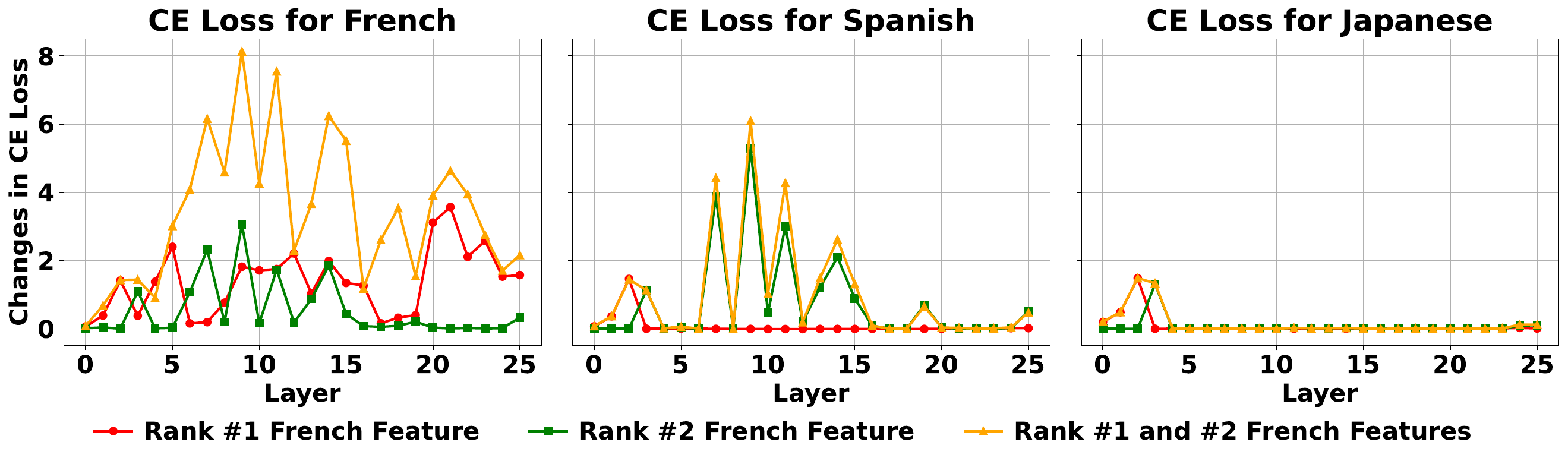}
  \caption{The change in CE loss for three languages after ablating French Features. Simultaneously ablating multiple French features exhibits a synergistic effect in French, while showing no synergistic effect on other languages. We provide results for Gemma 2 2B here, and additional results
can be found in Appendix~\ref{sec:multi_fea}.} 
  \label{fig:2_top_features}
\end{figure*}

In the previous section, we identified language-specific features that are closely related to monolingual texts. In this section, we examine how these language-specific features affect the language-specific capabilities of LLMs. Specifically, inspired by ~\citet{NEURIPS2024_f5454485,ferrando2024iknowentityknowledge}, we use \textit{directional ablation} to ``zero out'' language-specific features and observe the changes in the cross-entropy (CE) loss of texts in different languages within LLMs.
\subsection{Model Interventions}
\paragraph{Directional Ablation.} 
To analyze the impact of a feature $\mathbf{d}\in\mathbb{R}^{N}$ on the inference process of LLMs, \citet{NEURIPS2024_f5454485,ferrando2024iknowentityknowledge} introduce \textit{directional ablation} to ``zero out'' a feature in the residual stream activation $\mathbf{x} \in \mathbb{R}^N$. This is done by subtracting the projection of $\mathbf{x}$ onto $\mathbf{d}$ from $\mathbf{x}$:
\begin{equation}
    \mathbf{x}'\leftarrow\mathbf{x}-\hat{\mathbf{d}}\hat{\mathbf{d}}^\intercal\mathbf{x},\label{eq:ablation}
\end{equation}
where $\hat{\mathbf{d}}$ is the unit vector of $\mathbf{d}$. After obtaining the ablated residual stream, replace $\mathbf{x}$ with $\mathbf{x}'$ and continue the forward pass of the LLMs.

\subsection{Ablation of Language-Specific Features}
For each target language, layer, and LLM, we intervene in the inference process of LLMs using Eq.~\ref{eq:ablation} on the top-2 language-specific features of the target language. We then measure the changes in CE loss for both texts in the target language and texts in other languages after ablating language-specific features. The results are shown in Figure~\ref{fig:celoss}. We observe that: 
(1) Ablating language-specific features has a much larger impact on the CE loss of texts in the target language compared to texts in other languages.
(2) For different layers, the changes in CE loss of target language texts vary significantly.
These findings suggest that language-specific features play a crucial role in controlling the generation process for the target language. Ablating these features from the generation process of LLMs can lead to a loss of only specific language capabilities.

\subsection{Synergistic Language Features}
\label{sec:why_2_features}
We compare the CE loss for French, Spanish, and Japanese when using different numbers of French features for directional ablation. The results are shown in Figure~\ref{fig:2_top_features}. From this, we make the following observations:
(1) In some layers, simultaneously ablating the top 2 French features for French significantly impacts the CE loss more than ablating these features individually.
(2) In all layers, simultaneously ablating the top 2 French features for Spanish and Japanese results in a CE loss impact approximately equal to the sum of the effects when these features are ablated individually.
(3) The changes in CE loss for French are larger than those for Spanish and Japanese. The changes for Spanish are large in some layers, while for Japanese, they are nearly zero across all layers.

Based on these observations, we can conclude that for any target language, there exists a synergistic relationship among its features. Ablating multiple features simultaneously impacts significantly more than the sum of the effects when each feature is ablated individually. This synergistic effect is observed only when ablating language-specific features within its language.
Interestingly, in layers 7, 9, 10, 11, 14, and 15, the rank \#2 French feature is also among the top-2 Spanish features, explaining the significant changes in Spanish in some layers. 

\section{Enhancing Steering Vectors Using Language-Specific Features}
\label{sec:sv}
Having studied the basic characteristics of language-specific features, we now explore how to leverage these features in practice. Concretely, we use language-specific features as signals to guide steering vector~\citep{DBLP:journals/corr/abs-2308-10248,DBLP:conf/acl/RimskyGSTHT24,DBLP:journals/corr/abs-2411-08790}, in order to control the language in the model.
\subsection{Experimental Settings}
\paragraph{Tasks for Evaluation.}
We propose two tasks for evaluation. In the first task, \textit{Adversarial Language Identification}, given a text in language $A$, we prompt the model to identify its language. Our goal is to make the model identify the text as language $B$ instead. We use the CE loss for predicting language $B$ as the metric. In the second task,  \textit{Cross-Lingual Continuation}, given a text in language $A$, our goal is to make the model continue the text in language $B$. We use a language identification model from~\citet{DBLP:conf/acl/BurchellBBH23} to verify if the continuation is in language $B$. The success rate is used as the metric. Additionally, to measure the impact of the method on other language capabilities of LLMs, we also calculate the CE loss on Flores-10 without the original language when using the method.

\subsection{Methods}
\paragraph{Steering Vectors.}  
Steering vectors are vectors in the space of model activations that can guide a model's behavior when added to its internal activations~\citep{DBLP:journals/corr/abs-2308-10248,DBLP:conf/acl/RimskyGSTHT24,DBLP:journals/corr/abs-2411-08790}. To extract steering vectors for previously mentioned tasks, we use a subset in language $A$ from Flores-10 as positive prompt set, and another subset in language $B$ as negative prompt set, then we calculate the difference between the mean activations for positive and negative prompts at all token positions in layer $L$. This yields a steering vector $\mathbf{v}$, defined as:

\begin{equation}
    \mathbf{v} = \frac{1}{|\mathcal{X}_+|} \sum_{\mathbf{x} \in \mathcal{X}_+}  a_L(\mathbf{x}) - \frac{1}{|\mathcal{X}_-|} \sum_{\mathbf{x} \in \mathcal{X}_-}a_L(\mathbf{x}) ,\label{eq:sv}
\end{equation}
where $\mathcal{X}_+$ and $\mathcal{X}_-$ are positive and negative prompt sets, and  $a_\mathbf{L}(\mathbf{x})$ represent the mean activations in layer $L$ for prompt $\mathbf{x}$.

During inference, these steering vectors are directly added to the corresponding layer's activations across all tokens, replacing the original activations to continue the forward pass. By modifying the model's activations with the steering vector, the internal activation of the prompt can be steered from language $A$ towards language $B$, potentially improving performance on the language switching task.

\begin{figure}[t] 
\setlength{\abovecaptionskip}{0.10cm}
\setlength{\belowcaptionskip}{-0.30cm}
\centering
  \includegraphics[width=\columnwidth]{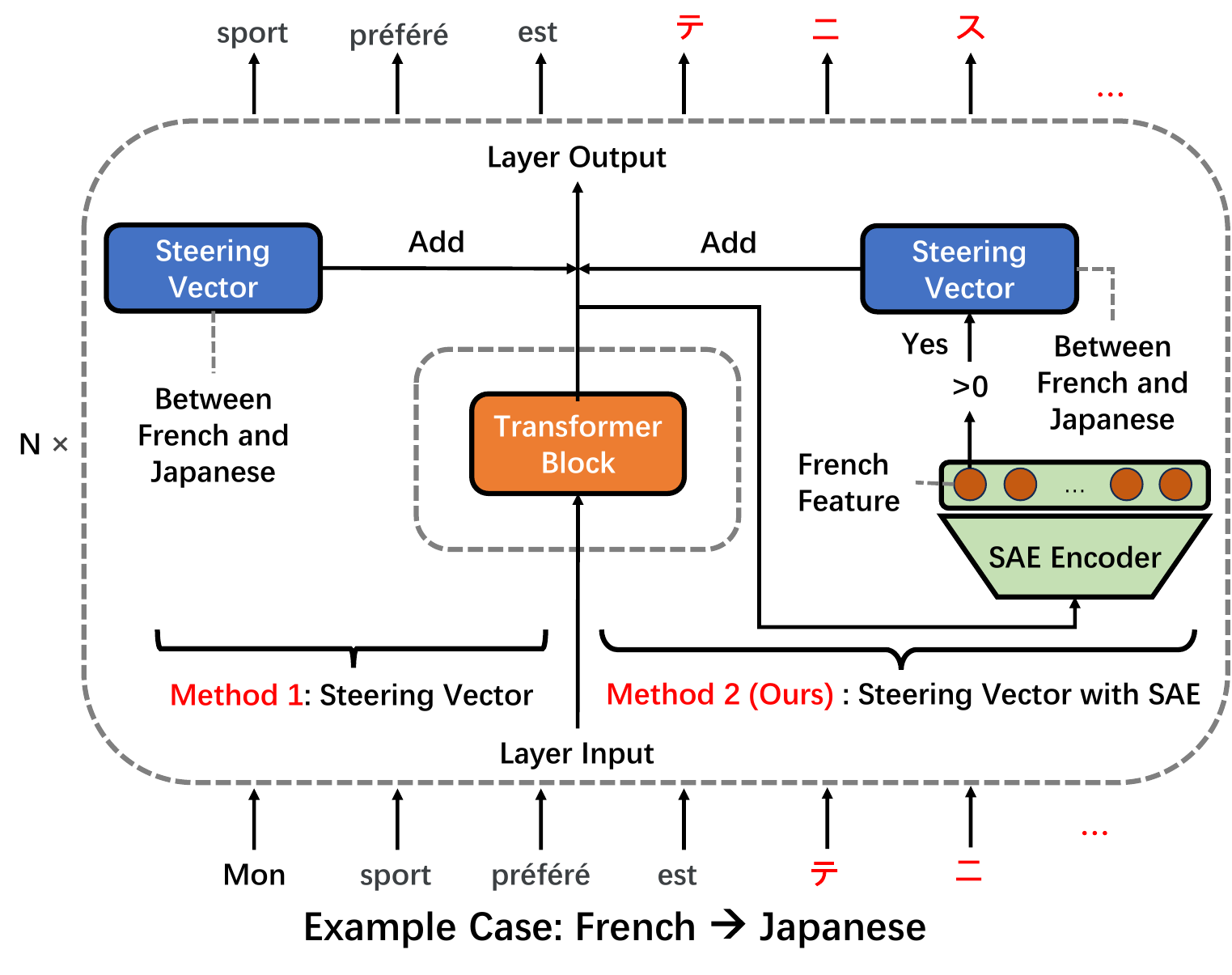}
  \caption{Comparison of steering vector and improved steering vector with SAE (ours): An example case from French to Japanese.} 
  \label{fig:pipline}
\end{figure}

\begin{figure}[t] 
\setlength{\abovecaptionskip}{0.10cm}
\setlength{\belowcaptionskip}{-0.40cm}
\centering
  \includegraphics[width=\columnwidth]{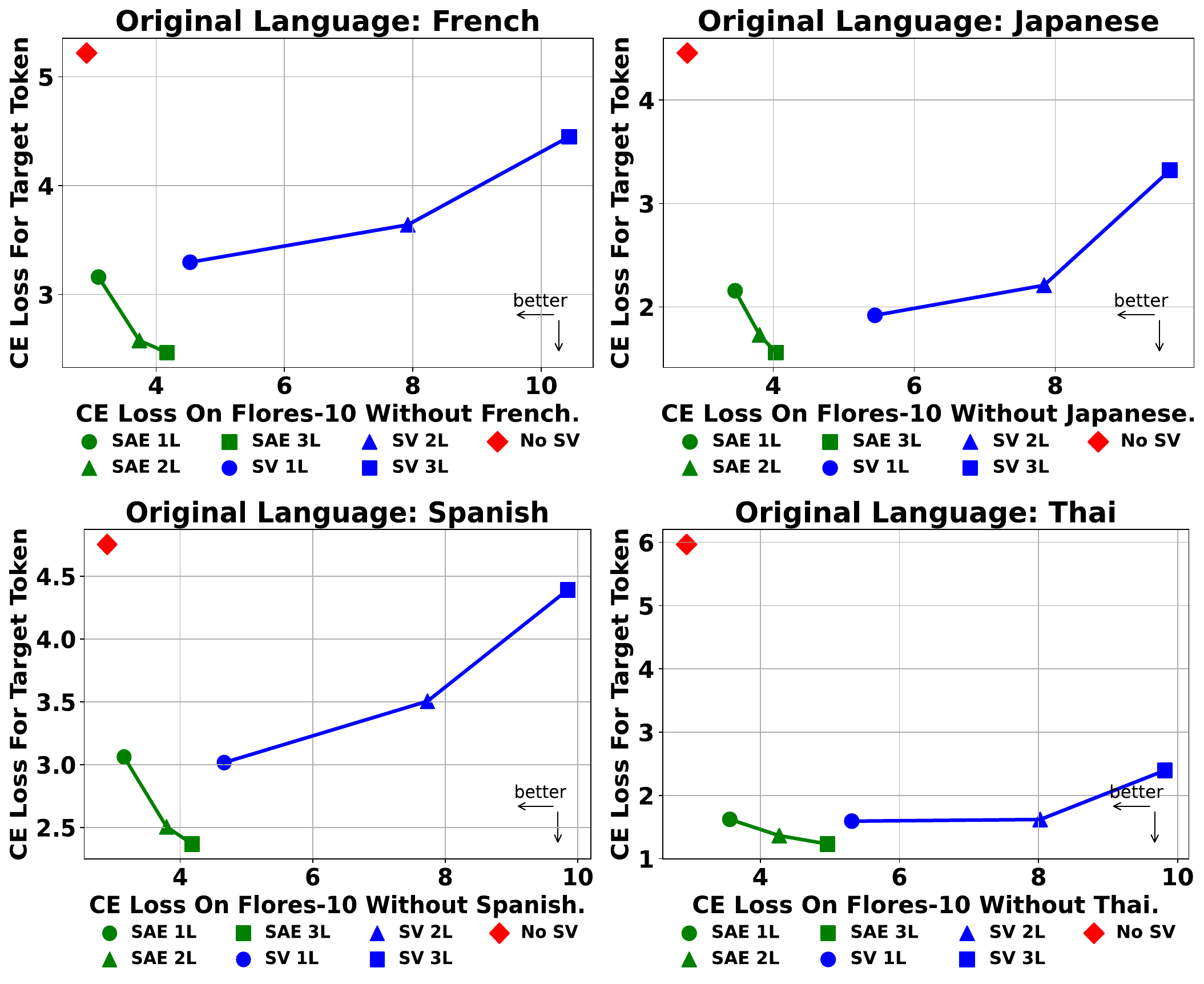}
  \caption{The y-axis shows the CE loss of LLMs identifying text in the original language as if it were in the target language, with lower values indicating better performance. The x-axis shows the impact on non-original language texts, with lower values indicating less impact. ``SV'' is the ``steering vector'' method, while ``SAE'' is our enhanced method. The suffix $k$L indicates results from modifying $k$ consecutive layers. Our method (green) provides a better balance between the metrics.
} 
  \label{fig:scatter}
\end{figure}

\begin{table*}[t]
\setlength{\abovecaptionskip}{0.10cm}
\setlength{\belowcaptionskip}{-0.40cm}
\centering
\setlength{\tabcolsep}{3mm}{
\begin{adjustbox}{max width=\textwidth}
\begin{tabular}{llccccccccc}
\toprule
\multirow{2}{*}{Model} & \multirow{2}{*}{Method} & \multicolumn{9}{c}{Success rate $ \uparrow $/ CE loss on other language $\downarrow$} \\
\cmidrule(lr){3-11} 
 &   & Es & Fr & Pt & Ja & Ko & Th & Vi & Zh& Ar\\
\midrule
\multirow{2}{*}{\text{Gemma 2 2B}} & \text{SV L1} & 92.1 / 4.7 & 92.6 / 4.5 & 84.2 / 4.7 & 86.1 / 5.4 & 95.2 / 5.3 & 85.7 / 5.3 & 91.1 / 4.6 & \textbf{84.7} / 5.2 & \textbf{88.3} / 4.6  \\
& \text{SAE L3} & \textbf{95.8} / \textbf{4.2} & \textbf{96.7} / \textbf{4.2} & \textbf{84.4} / \textbf{4.4} & \textbf{89.2} / \textbf{4.0} & \textbf{95.4} / \textbf{4.4} & \textbf{90.7} / \textbf{5.0} & \textbf{91.3} / \textbf{3.4} & 71.9 / \textbf{4.3} & 81.3 / \textbf{3.9}  \\
\midrule
\multirow{2}{*}{\text{Gemma 2 9B}} & \text{SV L1} & 82.2 / 4.1 & 85.3 / 4.0 & 76.4 / 4.1 & 83.4 / 4.5 & 93.0 / 4.6 & \textbf{88.7} / \textbf{4.6} & 83.6 / 4.1 & \textbf{79.5} / 4.4 & \textbf{84.2} / 4.1  \\
& \text{SAE L3} & \textbf{96.2} / \textbf{3.4} & \textbf{94.6} / \textbf{2.9} & \textbf{86.1} / \textbf{3.2} & \textbf{86.3} / \textbf{3.0} & \textbf{93.6} / \textbf{4.3} & 85.6 / 4.9 & \textbf{95.3} / \textbf{2.8} & 77.0 / \textbf{3.2} & 78.3 / \textbf{4.0}  \\
\midrule
\multirow{2}{*}{\text{Llama-3.1-8B}} & \text{SV L1} & 85.7 / 3.7 & 86.8 / 3.5 & 79.7 / 3.6 & 79.1 / 4.4 & 90.0 / 4.3 & \textbf{88.4} / 4.4 & 85.0 / \textbf{3.6} & 77.0 / 4.1 & 85.2 / 3.8  \\
& \text{SAE L3} & \textbf{97.0} / \textbf{3.0} & \textbf{96.1} / \textbf{2.7} & \textbf{80.0} / \textbf{2.7} & \textbf{86.6} / \textbf{3.6} & \textbf{95.2} / \textbf{3.2} & 78.0 / \textbf{3.3} & \textbf{94.8} / 4.7 & \textbf{91.2} / \textbf{3.7} & \textbf{88.4} / \textbf{2.7}  \\

\bottomrule
\end{tabular}
\end{adjustbox}
}
\caption{The results of \textit{Cross-Lingual Continuation} task. Our SAE method can surpass the SV method across both metrics in most cases and achieve a much better balance between the metrics. 
\label{tab:lid_results}}
\end{table*}

\paragraph{Improved Steering Vectors with Language-Specific Features.} 
There are two main drawbacks of steering vectors:
(1) Adding a steering vector across all tokens, including non-target language tokens, leads to an increase in CE loss in non-target language texts.
(2) Using a steering vector in multiple layers simultaneously does not lead to better performance since frequent adjustments may drastically change the normal distribution of activation values, as demonstrated in Figure~\ref{fig:scatter}.
As a result, we propose using the activation of language-specific features as a signal to determine whether to use a steering vector for a token. Concretely, for a steering vector from language ${B}$ to language ${A}$ in layer ${L}$ and a model activation ${x}$ in layer ${L}$, if the activation of the top-2 language ${B}$ features of $\mathbf{x}$ are non-zero, we add the steering vector to it and continue the forward pass.
The rationale for employing language-specific features is twofold: First, it ensures that steering vectors affect only target language tokens, preventing increased CE loss in non-target language texts. Second, by applying steering vectors selectively based on specific language feature activations, it minimizes excessive adjustments across layers, maintaining activation value distribution and enhancing model stability.\footnote{In cases where there is no ambiguity, we abbreviate the ``steering vector'' as ``SV'' and our improved ``steering vector with SAE'' as ``SAE.'' And we add a suffix $k$L to indicate modification of $k$ consecutive layers.}

\subsection{Results}
\paragraph{Better Performance on Adversarial Language Identification.}
In Figure~\ref{fig:scatter}, we present the results for  \textit{Adversarial Language Identification}. The experiments cover the steering vector both with and without SAE across one, two, and three consecutive layers, denoted as ${1L}$, ${2L}$, and ${3L}$. From the figure, we make the following observations: 
(1) As the number of modified layers increases, our SAE method achieves better performance in  \textit{Adversarial Language Identification}, whereas the performance of the SV method decreases. This indicates that SAE is more effective at  \textit{Adversarial Language Identification }and more robust when applied to multiple layers.
(2) As the number of modified layers increases, the CE loss on other languages also increases. However, the rate of increase with our SAE method is much smaller than with the SV method. In most cases, the CE loss on other languages for SAE applied to three consecutive layers is even lower than that for SV applied to a single layer. These results suggest that our SAE method achieves a better balance between the two metrics on \textit{Adversarial Language Identification}.

\paragraph{Better Performance on Cross-Lingual Continuation.}
As illustrated in Figure~\ref{fig:scatter}, the performance of the SV method declines rapidly; hence, we only report the results of SV 1L for \textit{Cross-Lingual Continuation}. The results are shown in Table~\ref{tab:lid_results}, where we observe the following:
(1) For different languages and models, SAE 3L outperforms SV 1L in both success rate and CE loss in most cases.
(2) In some cases, SAE 3L achieves a better CE loss but with a lower success rate compared to SV 1L. Since these two metrics are generally a trade-off, this does not imply that the SAE method is inferior to the SV method.
These results suggest that our SAE method can surpass the SV method across both metrics in most cases. 

\section{Related Works} 
\paragraph{Multilingual Mechanism of LLMs}
Multilingual mechanisms of LLMs are mainly studied through neuron-based and ``logit lens''~\citep{nostalgebraist2020logitlens} methods.
Neuron-based aim to identify language-specific neurons within LLMs and modify these neurons to assess their impact on the corresponding language~\citep{DBLP:conf/acl/ZhangZ0G024,DBLP:conf/nips/0006ZCKB24,DBLP:conf/acl/TangLH0WZWW24,DBLP:conf/naacl/KojimaOIYM24}. For example,~\citet{DBLP:conf/acl/ZhangZ0G024} discover that removing certain neurons in LLMs leads to a significant performance decrease in some languages.~\citet{DBLP:conf/nips/0006ZCKB24} introduce PLND to identify activated neurons for inputs in different languages, and hypothesize that in the intermediate layers, LLMs employ English for thinking. Moreover,~\citet{DBLP:conf/acl/TangLH0WZWW24,DBLP:conf/naacl/KojimaOIYM24} explore methods to identify language-specific neurons within LLMs.
However, these methods can be complex and unreliable due to ``superposition,''~\citep{elhage2022toymodelssuperposition} where multiple concepts can be encoded in a single neuron. ``Logit lens'' methods derive token distributions from intermediate layers using the output layer's unembedding matrix.~\citet{DBLP:conf/acl/WendlerVM024} find that Llama2~\citep{touvron2023llama2openfoundation} might use English as an internal language in intermediate layers, and~\citet{zhong2024englishcentricllmslanguagemultilingual} extend the conclusion, showing that LLMs with continued pre-training in Japanese employ both Japanese and English in intermediate layers. Due to the varying distribution of residual streams across different layers, the ``logit lens'' method often has significant errors except in the last few layers, making the analysis sometimes unreliable.
\paragraph{SAEs}
SAEs are a specialized form of autoencoders designed to decompose language model activations into a linear combination of SAE feature directions~\citep{bricken2023monosemanticity,cunningham2023sparse}. Typically, the activations of neurons in deep neural networks do not have a straightforward, human-understandable interpretation. However, SAEs can transform these activations into a higher-dimensional latent space, which is potentially more interpretable. For instance,~\citet{cunningham2023sparse} identify features associated with apostrophes. Meanwhile,~\citet{ferrando2025iknowentityknowledge} discover features that indicate whether LLMs recognize a particular entity. Furthermore,~\citet{paulo2024automaticallyinterpretingmillionsfeatures} develop an open-source automated pipeline to generate and evaluate natural language explanations for SAE features using LLMs. Their work confirms that SAEs are indeed significantly more interpretable than individual neurons.

\section{Conlusion}
In this study, we explored the underlying mechanisms of multilingual capabilities in LLMs using SAEs to achieve a more refined analysis.
By introducing a novel metric for monolinguality, we found that certain features were strongly tied to specific languages. And directional ablation confirmed the significant role these features play in enhancing language-specific capabilities, with combined feature ablation yielding greater improvements than individual ablation. Additionally, we improved steering vectors using these SAE-derived features, achieving better performance and robustness. 
Building upon the insights gained from this work, an exciting avenue for future research is to utilize these language-specific features to guide the training process of multilingual language models. 
\section*{Limitations}
This study has several limitations that we plan to address in the future.
First, although our method performs well across 10 different languages, it does not yet cover certain low-resource languages. Investigating these underrepresented languages will enhance our analysis.
Second, while our SAE-based steering vectors outperform the original steering vectors in most cases (see Table~\ref{tab:lid_results}), there are instances where our method falls short. Therefore, exploring a more refined approach to improve performance is worthwhile.
Third, the SAEs used in our experiments were not trained on curated multilingual data. It would be advantageous to train SAEs using high-quality multilingual datasets.

\section*{Acknowledgements}
This research was also supported by the advanced computing resources provided by the Supercomputing Center of the USTC.
\bibliography{custom}

\appendix

\newpage
\section{Implementation Details}
For each experiment, we don't perform any sampling during generation to avoid randomness. For SAEs from Gemma Scope~\citep{lieberum2024gemmas}, we choose the one with the second smallest $L_0$ value for each layer. For SAEs from Llama Scope~\citep{he2024llamas}, we use the model available at \url{https://huggingface.co/fnlp/Llama3_1-8B-Base-LXR-8x/tree/main}.
\section{Flores-10}
We extract a subset called Flores-10 from Flores-200, which includes 10 languages: English (en), Spanish (es), French (fr), Japanese (ja), Korean (ko), Portuguese (pt), Thai (th), Vietnamese (vi), Chinese (zh), and Arabic (ar). In Section~\ref{sec:lan_specific_feature}, we use the first 100 data points in the dev set for each language to identify language-specific features. In Section~\ref{sec:ablate}, we use the first 100 data points in the dev set for each language to generate the steering vector, and perform experiments using 500 data points in the dev set for each language that do not overlap with the first 100 data points.

\section{Additional Results of $\nu$ }
\label{Appendix:nu}
\begin{figure*}[t] 
\centering
  \includegraphics[width=\textwidth]{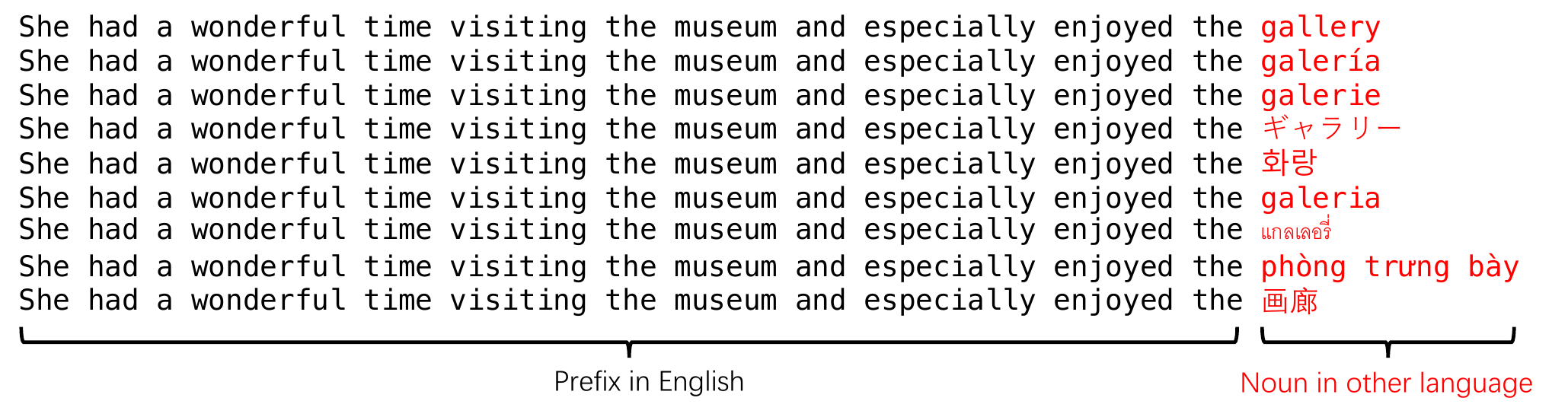}
  \caption{Example data of our code-switching dataset.}
  \label{fig:cs_data}
\end{figure*}

\begin{figure*}[t] 
\centering
  \includegraphics[width=0.75\textwidth]{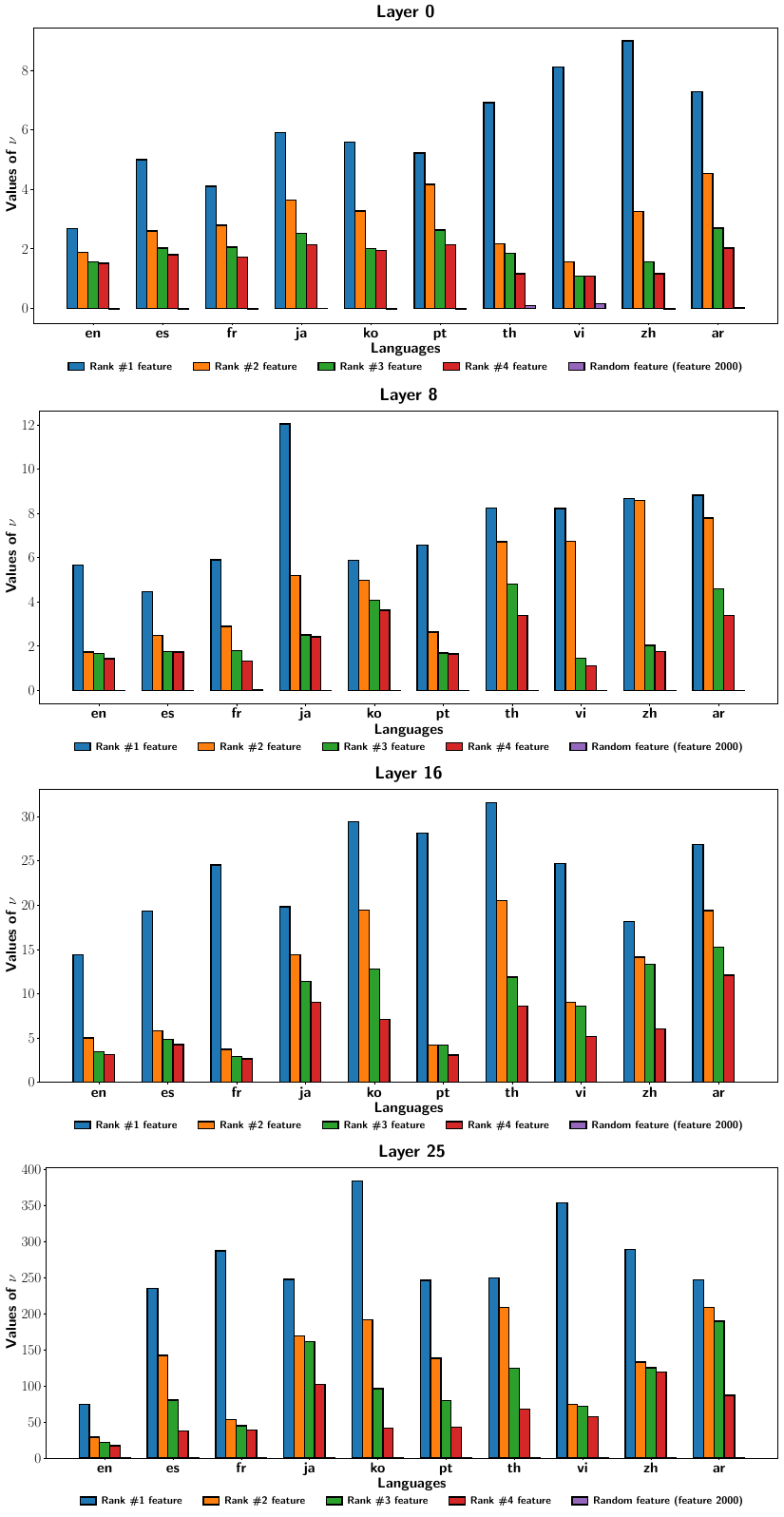}
  \caption{The values of $\nu$ of Gemma 2 2B.} 
  \label{fig:gemma2nu}
\end{figure*}

\begin{figure*}[t] 
\centering
  \includegraphics[width=0.75\textwidth]{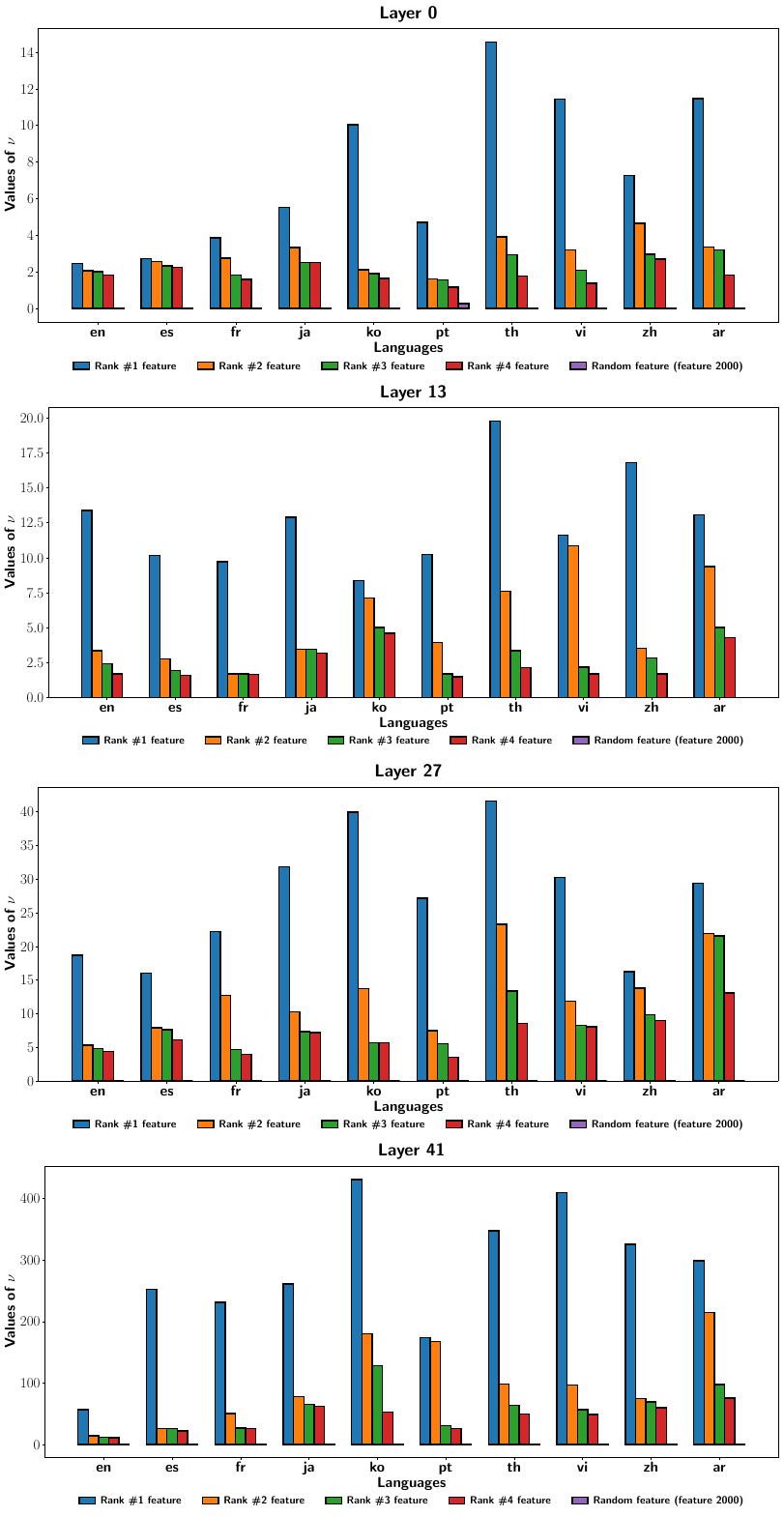}
  \caption{The values of $\nu$ of Gemma 2 9B.} 
  \label{fig:gemma9nu}
\end{figure*}

\begin{figure*}[t] 
\centering
  \includegraphics[width=0.75\textwidth]{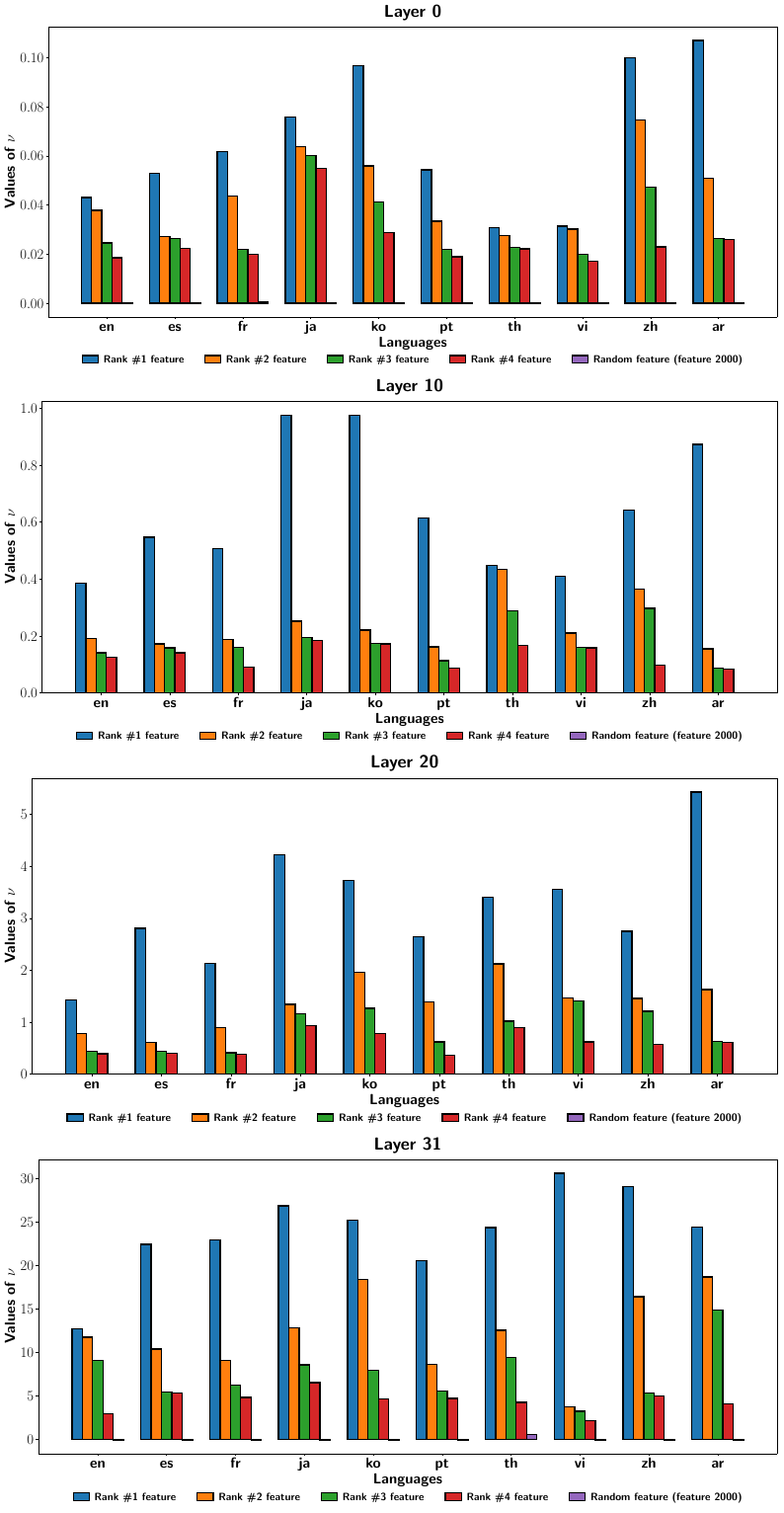}
  \caption{The values of $\nu$ of Llama-3.1-8B.} 
  \label{fig:llamanu}
\end{figure*}

Additional results of $\nu$ for 3 different LLMs are demonstrated in Figure~\ref{fig:gemma2nu}-\ref{fig:llamanu}. We report the results at four different levels: ``first layers'', ``$\frac{1}{3}$ of the total layers'', ``$\frac{2}{3}$ of the total layers'', and the ``final layer''. Similarly to the results in Figure~\ref{fig:histogram_gemma-2-2b_layer_20}, the top-ranked features possess strong monolingual characteristics, and in most scenarios, the top-1 feature suffices in capturing these characteristics.

\section{Additional Results for Directional Ablation}
\label{Appendix:Directional_Ablation}
\begin{figure*}[t] 
\centering
  \includegraphics[width=0.75\textwidth]{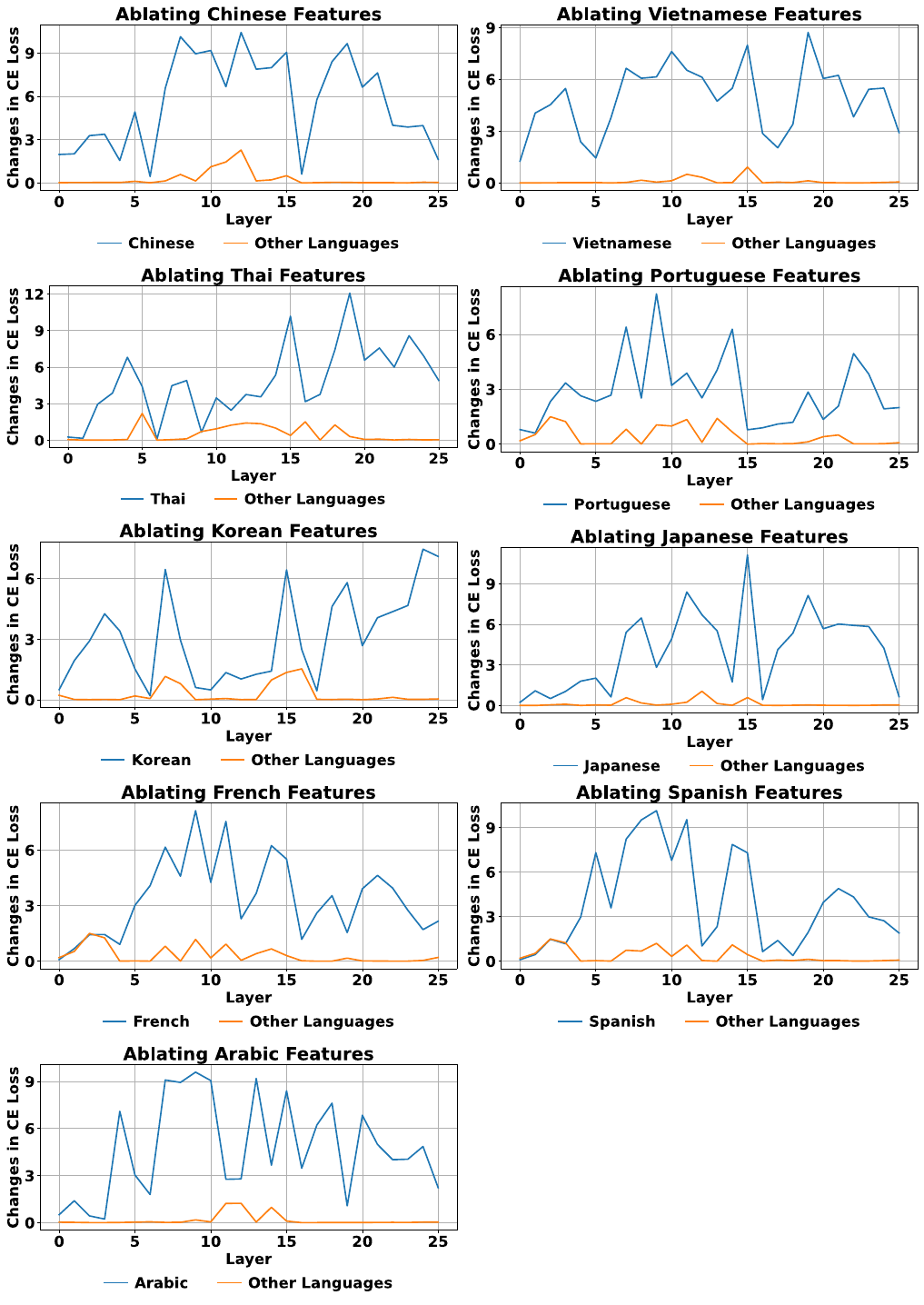}
  \caption{The changes in CE loss on texts in the target language and texts in other languages after ablating language-specific features for Gemma 2 2B.} 
  \label{fig:gemma_celoss}
\end{figure*}

\begin{figure*}[t] 
\centering
  \includegraphics[width=0.75\textwidth]{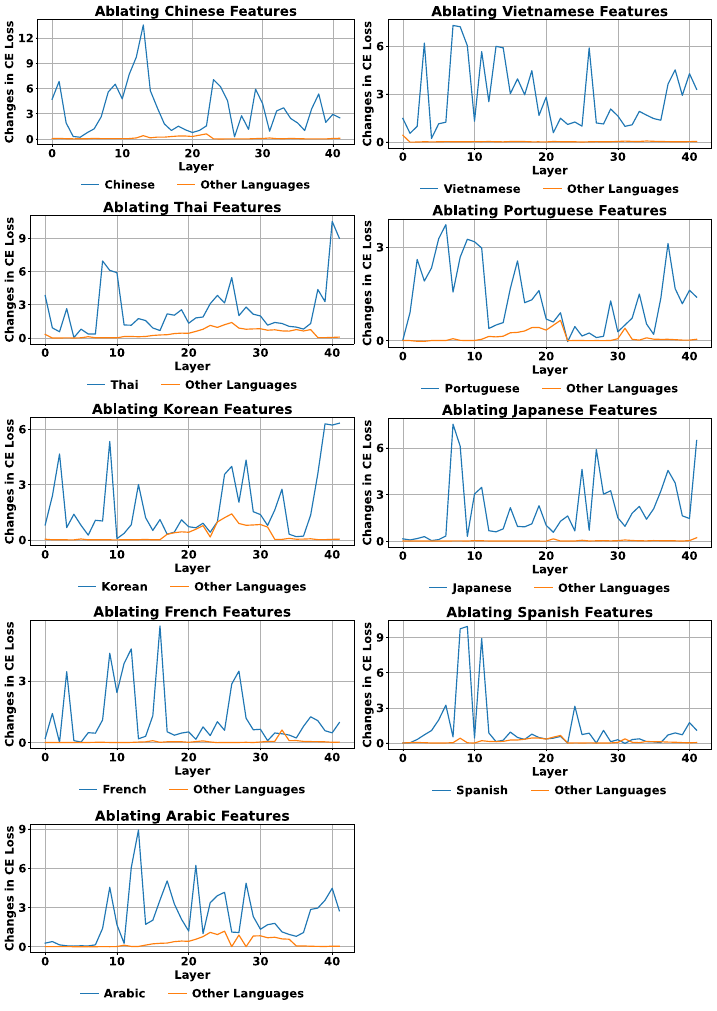}
  \caption{The changes in CE loss on texts in the target language and texts in other languages after ablating language-specific features for Gemma 2 9B.} 
  \label{fig:gemma9_celoss}
\end{figure*}

\begin{figure*}[t] 
\centering
  \includegraphics[width=0.75\textwidth]{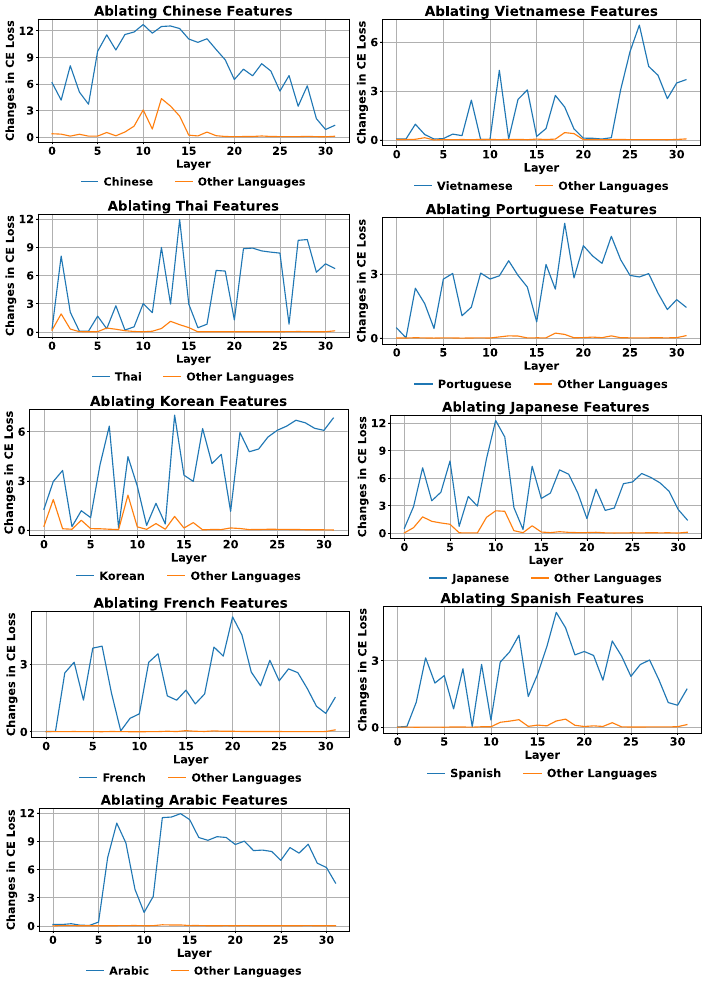}
  \caption{The changes in CE loss on texts in the target language and texts in other languages after ablating language-specific features for Llama-3.1-8B.} 
  \label{fig:llama_celoss}
\end{figure*}

Additional results for directional ablation for 3 different LLMs are demonstrated in Figure~\ref{fig:gemma_celoss}-\ref{fig:llama_celoss}. The results are similar to those in Figure~\ref{fig:celoss}, where ablating language-specific features has a much larger impact on the CE loss of texts in the target language compared to texts in other languages.

\section{Additional Results for Code-Switching}
\label{sec:cs_appendix}
\begin{figure*}[t] 
\centering
  \includegraphics[width=\textwidth]{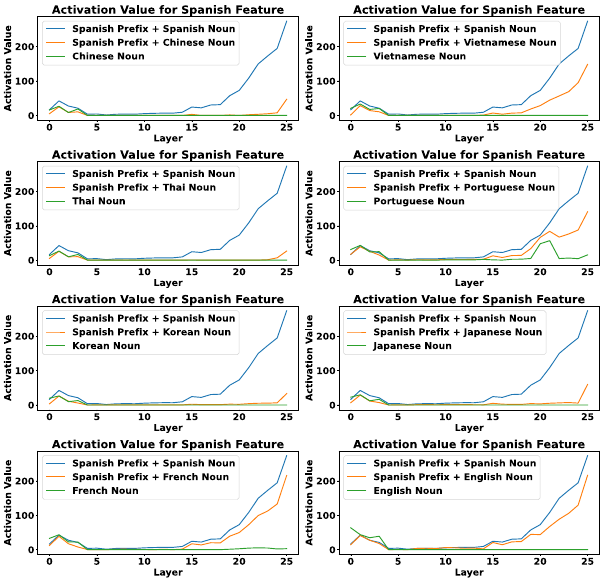}
  \caption{The mean activation values for the Spanish feature with various noun and prefix combinations for Gemma 2 2B.}
  \label{fig:es_cs_gemma2}
\end{figure*}

\begin{figure*}[t] 
\centering
  \includegraphics[width=\textwidth]{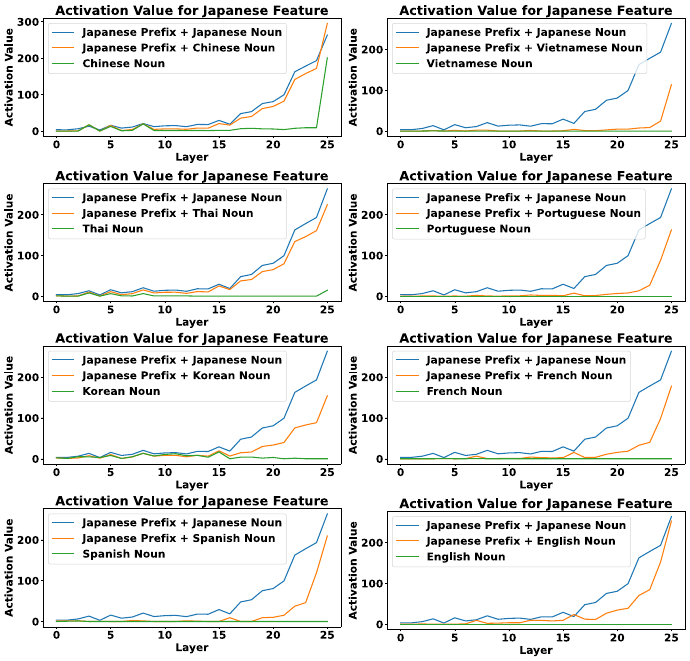}
  \caption{The mean activation values for the Japanese feature with various noun and prefix combinations for Gemma 2 2B.}
  \label{fig:ja_cs_gemma2}
\end{figure*}

\begin{figure*}[t] 
\centering
  \includegraphics[width=\textwidth]{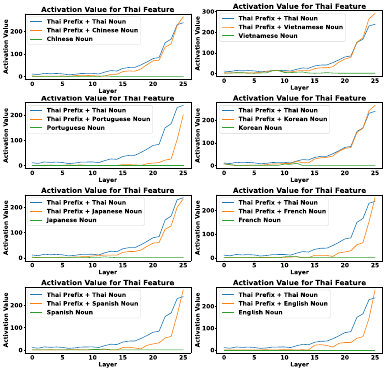}
  \caption{The mean activation values for the Thai feature with various noun and prefix combinations for Gemma 2 2B.}
  \label{fig:th_cs_gemma2}
\end{figure*}

\begin{figure*}[t] 
\centering
  \includegraphics[width=\textwidth]{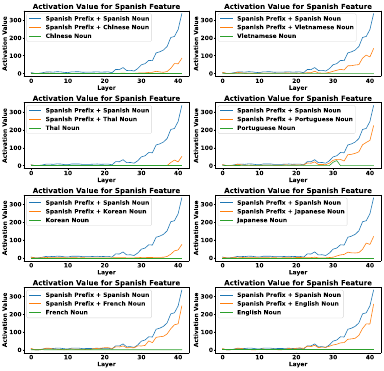}
  \caption{The mean activation values for the Spanish feature with various noun and prefix combinations for Gemma 2 9B.}
  \label{fig:es_cs_gemma9}
\end{figure*}

\begin{figure*}[t] 
\centering
  \includegraphics[width=\textwidth]{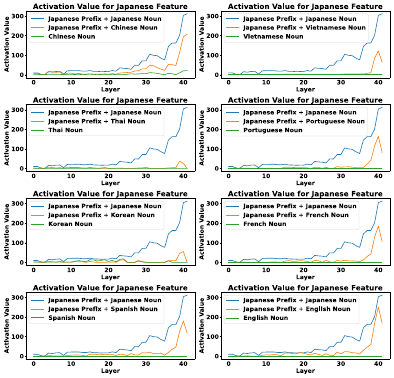}
  \caption{The mean activation values for the Japanese feature with various noun and prefix combinations for Gemma 2 9B.}
  \label{fig:ja_cs_gemma9}
\end{figure*}

\begin{figure*}[t] 
\centering
  \includegraphics[width=\textwidth]{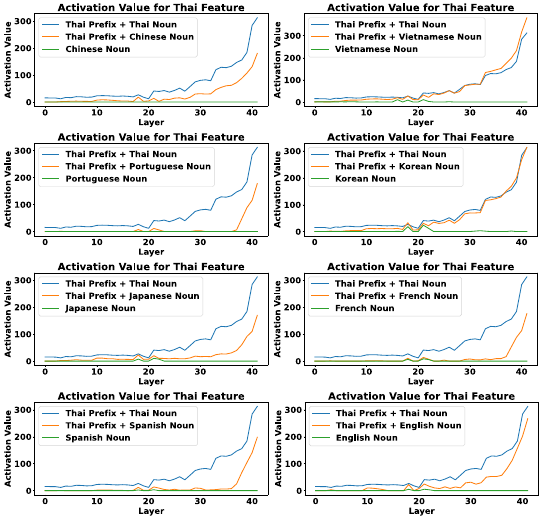}
  \caption{The mean activation values for the Thai feature with various noun and prefix combinations for Gemma 2 9B.}
  \label{fig:th_cs_gemma9}
\end{figure*}

\begin{figure*}[t] 
\centering
  \includegraphics[width=\textwidth]{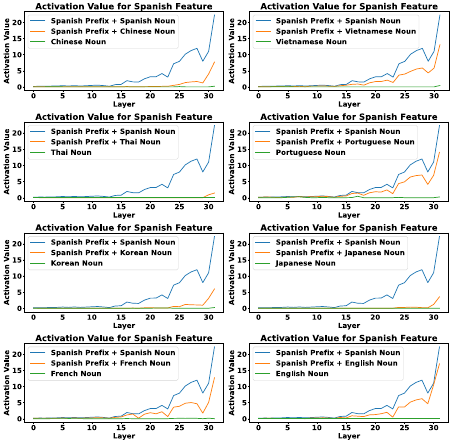}
  \caption{The mean activation values for the Spanish feature with various noun and prefix combinations for Llama-3.1-8B.}
  \label{fig:es_cs_llama}
\end{figure*}

\begin{figure*}[t] 
\centering
  \includegraphics[width=\textwidth]{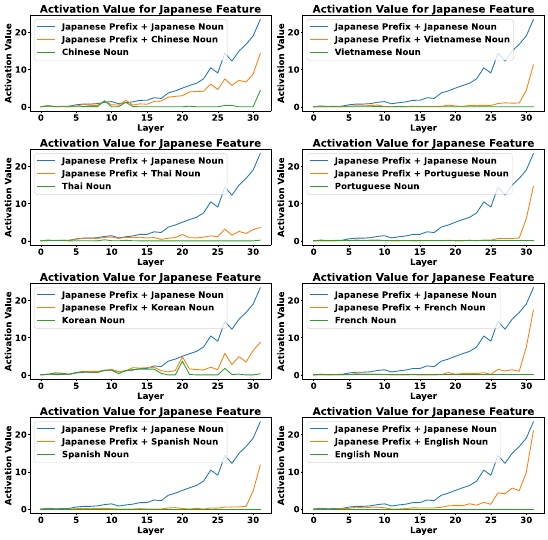}
  \caption{The mean activation values for the Japanese feature with various noun and prefix combinations for Llama-3.1-8B.}
  \label{fig:ja_cs_llama}
\end{figure*}

\begin{figure*}[t] 
\centering
  \includegraphics[width=\textwidth]{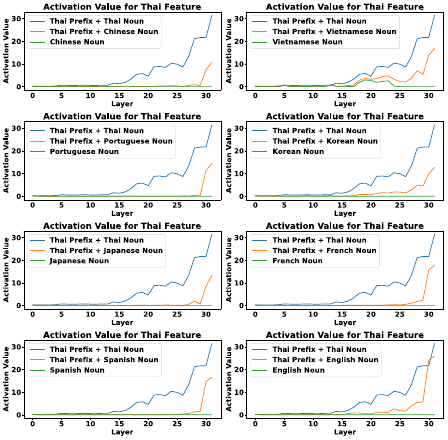}
  \caption{The mean activation values for the Thai feature with various noun and prefix combinations for Llama-3.1-8B.}
  \label{fig:th_cs_llama}
\end{figure*}

\begin{figure*}[t] 
\centering
  \includegraphics[width=\textwidth]{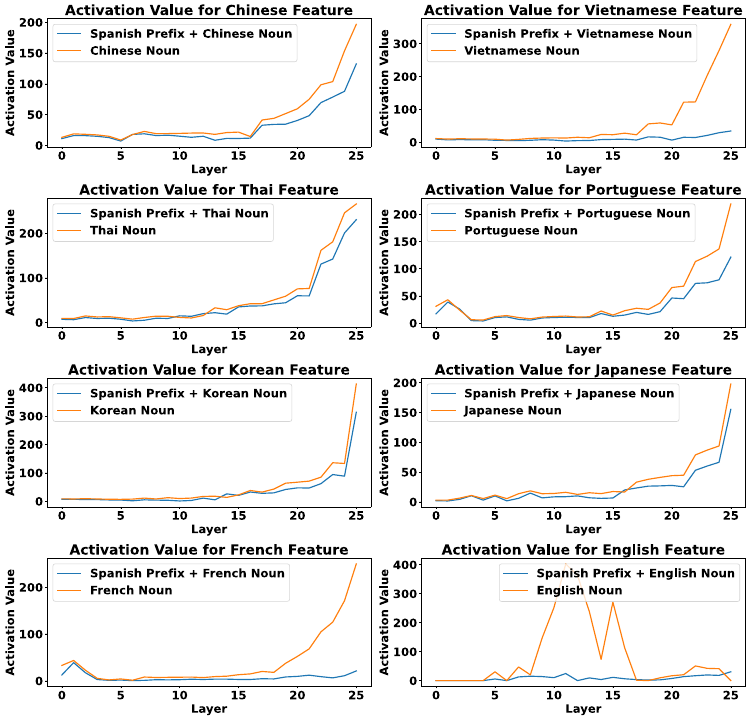}
  \caption{The mean activation values for various features with Spanish prefix for Gemma 2 2B.}
  \label{fig:es_cs_gemma2_de}
\end{figure*}

\begin{figure*}[t] 
\centering
  \includegraphics[width=\textwidth]{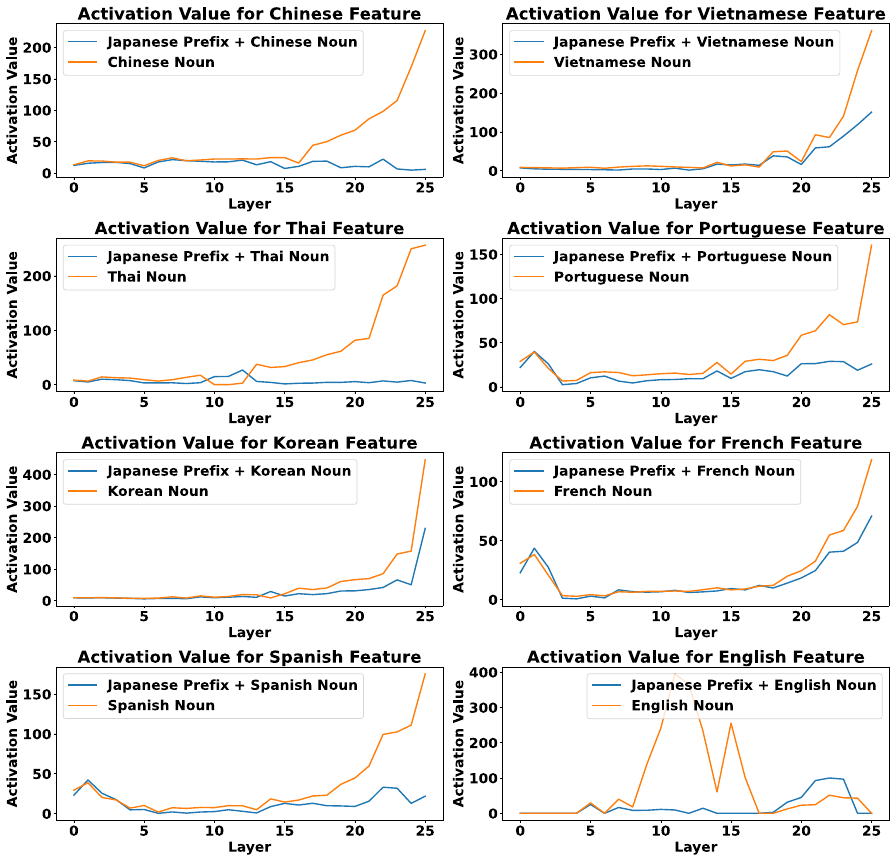}
  \caption{The mean activation values for various features with Japanese prefix for Gemma 2 2B.}
  \label{fig:ja_cs_gemma2_de}
\end{figure*}

\begin{figure*}[t] 
\centering
  \includegraphics[width=\textwidth]{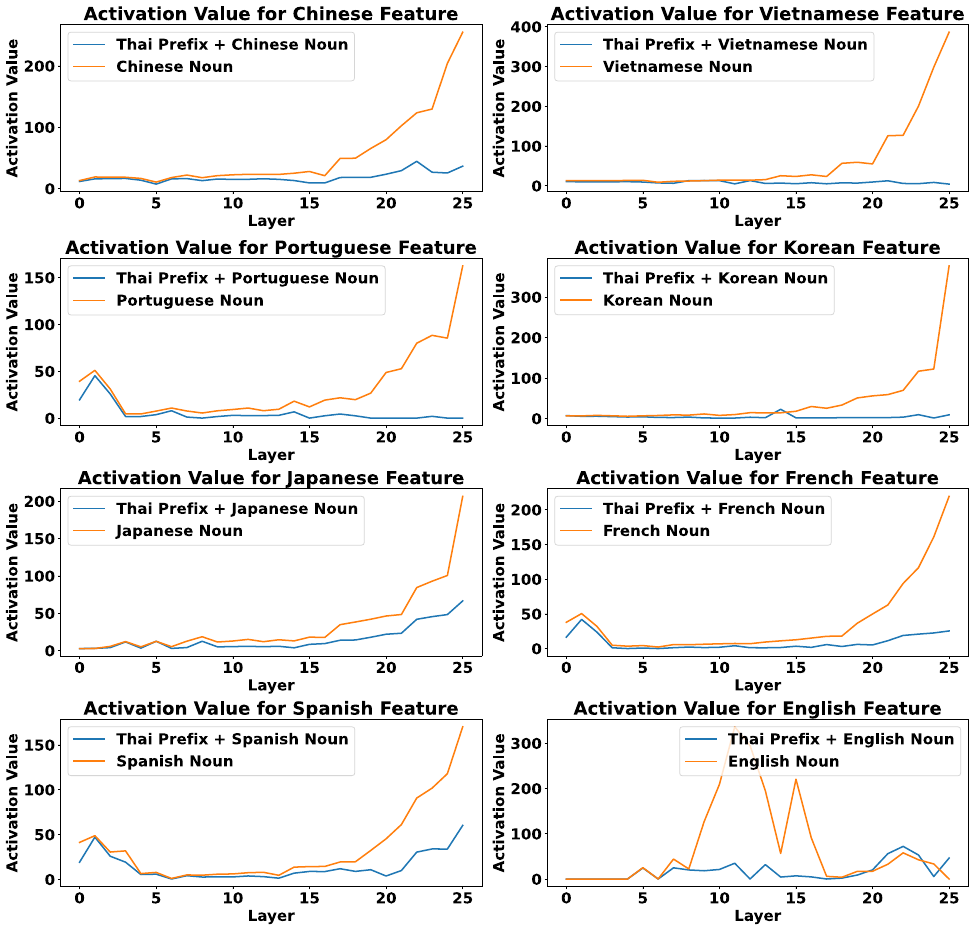}
  \caption{The mean activation values for various features with Thai prefix for Gemma 2 2B.}
  \label{fig:th_cs_gemma2_de}
\end{figure*}

\begin{figure*}[t] 
\centering
  \includegraphics[width=\textwidth]{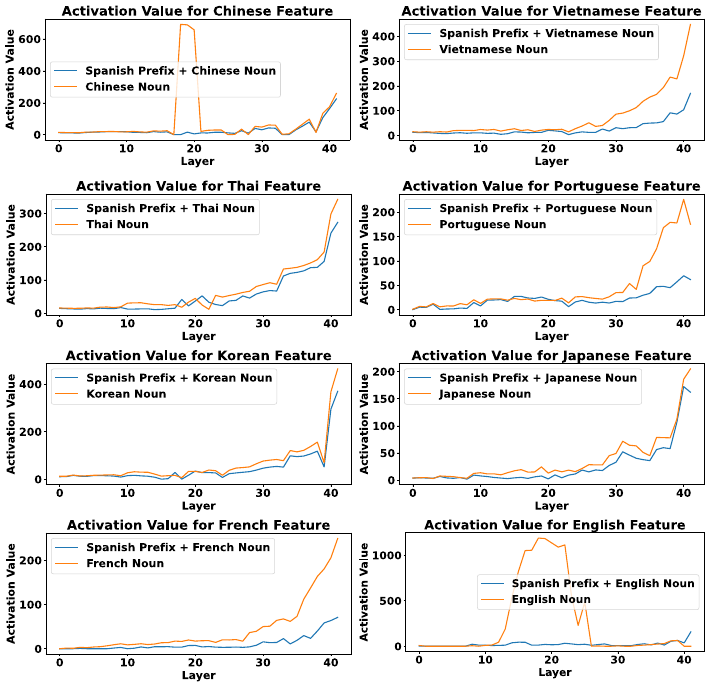}
  \caption{The mean activation values for various features with Spanish prefix for Gemma 2 9B.}
  \label{fig:es_cs_gemma9_de}
\end{figure*}

\begin{figure*}[t] 
\centering
  \includegraphics[width=\textwidth]{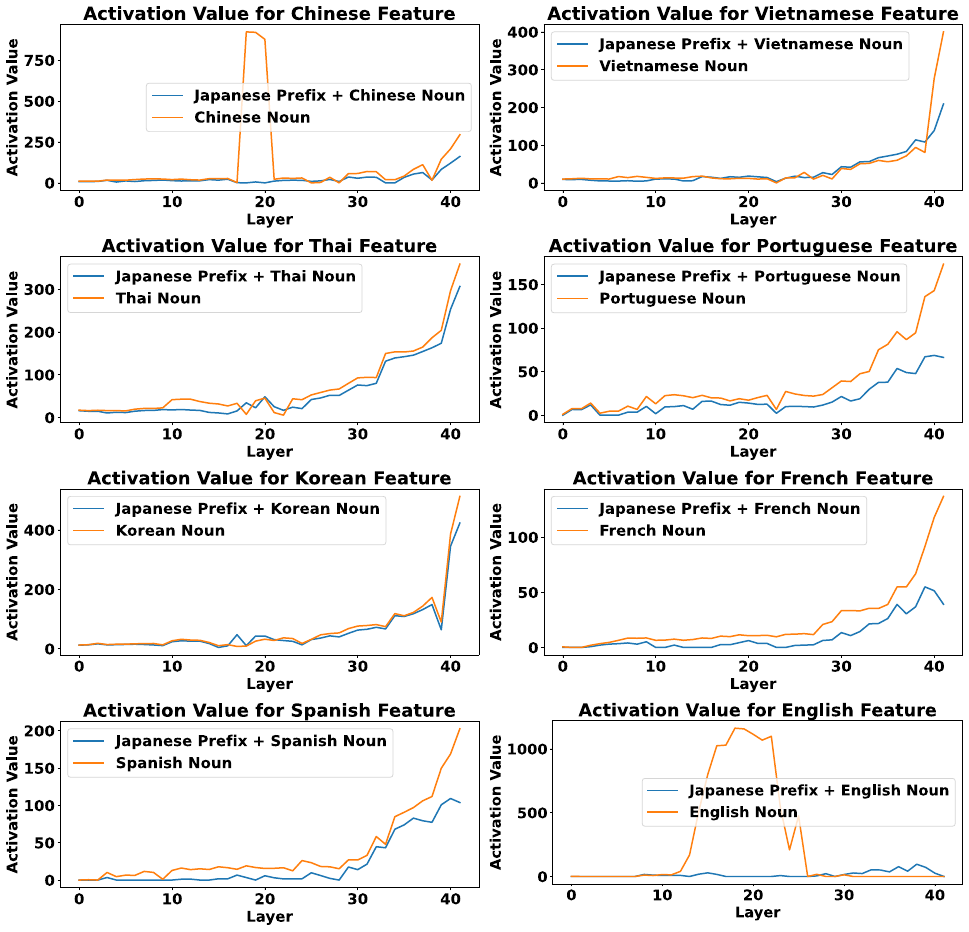}
  \caption{The mean activation values for various features with Japanese prefix for Gemma 2 9B.}
  \label{fig:ja_cs_gemma9_de}
\end{figure*}

\begin{figure*}[t] 
\centering
  \includegraphics[width=\textwidth]{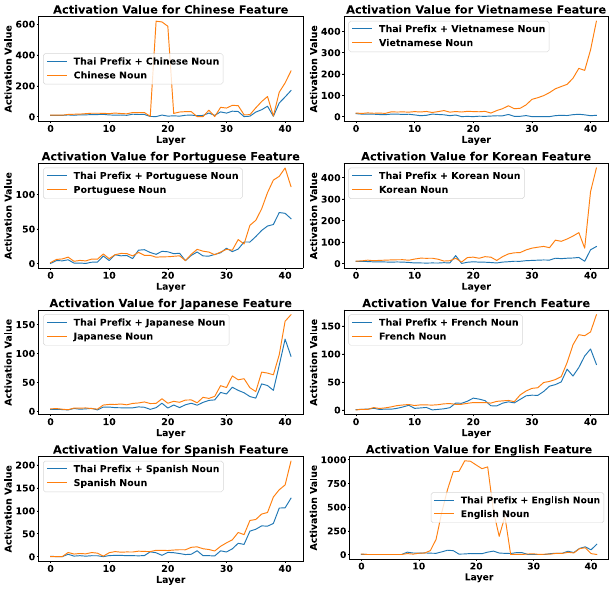}
  \caption{The mean activation values for various features with Thai prefix for Gemma 2 9B.}
  \label{fig:th_cs_gemma9_de}
\end{figure*}

\begin{figure*}[t] 
\centering
  \includegraphics[width=\textwidth]{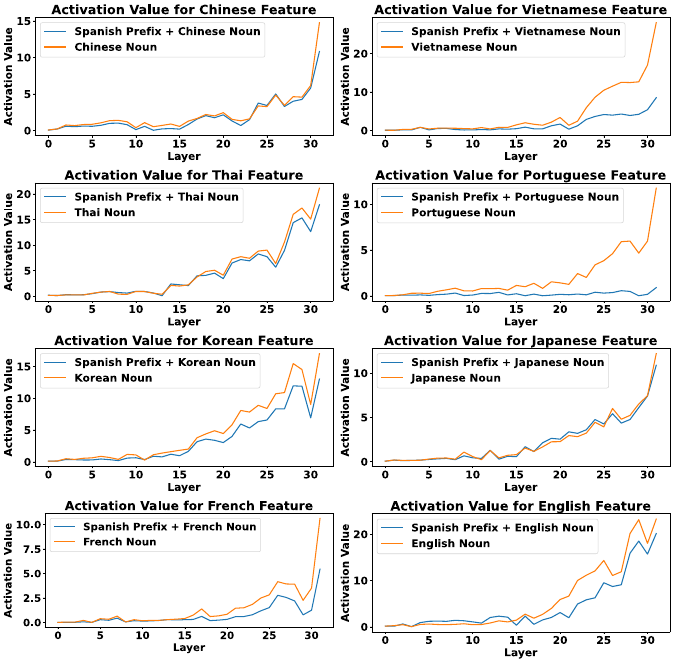}
  \caption{The mean activation values for various features with Spanish prefix for Llama-3.1-8B.}
  \label{fig:es_cs_llama_de}
\end{figure*}

\begin{figure*}[t] 
\centering
  \includegraphics[width=\textwidth]{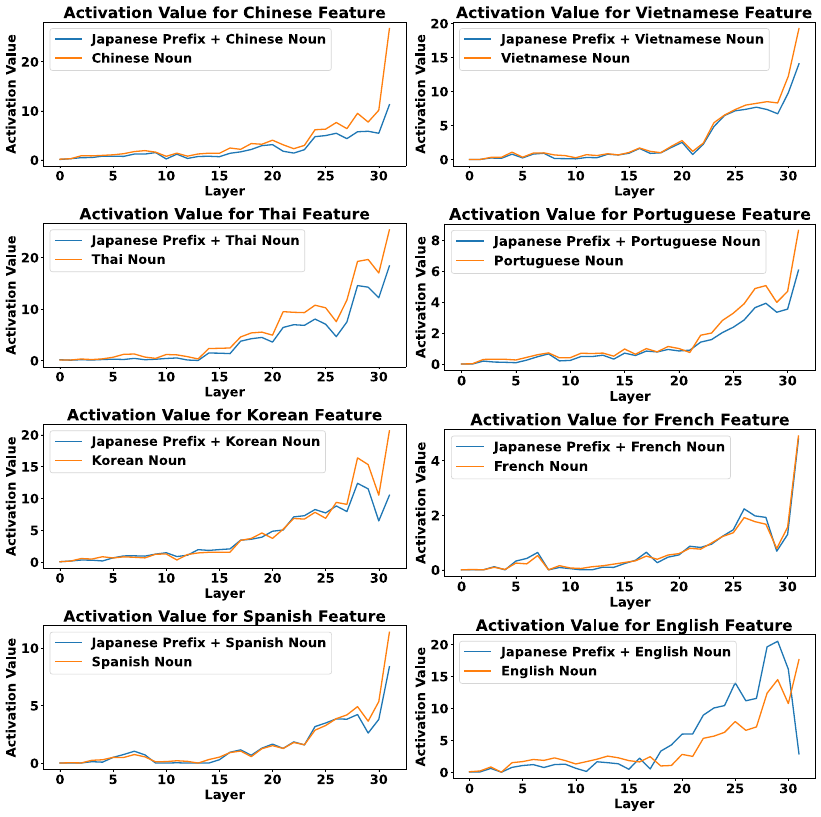}
  \caption{The mean activation values for various features with Japanese prefix for Llama-3.1-8B.}
  \label{fig:ja_cs_llama_de}
\end{figure*}

\begin{figure*}[t] 
\centering
  \includegraphics[width=\textwidth]{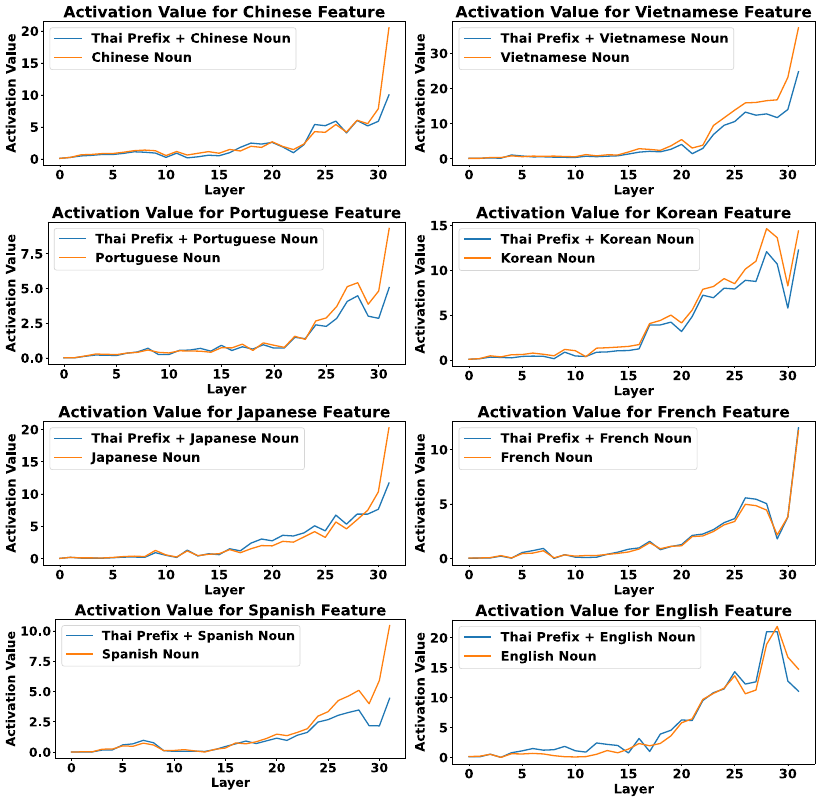}
  \caption{The mean activation values for various features with Thai prefix for Llama-3.1-8B.}
  \label{fig:th_cs_llama_de}
\end{figure*}

Additional results for code-switching are demonstrated in Figure~\ref{fig:es_cs_gemma2}-\ref{fig:th_cs_llama_de}. The results are similar to those in Figure~\ref{fig:codeswitch} and~\ref{fig:codeswitch2}.

\section{Additional Results for Multiple Features}
\label{sec:multi_fea}
\begin{figure*}[t] 
\centering
  \includegraphics[width=\textwidth]{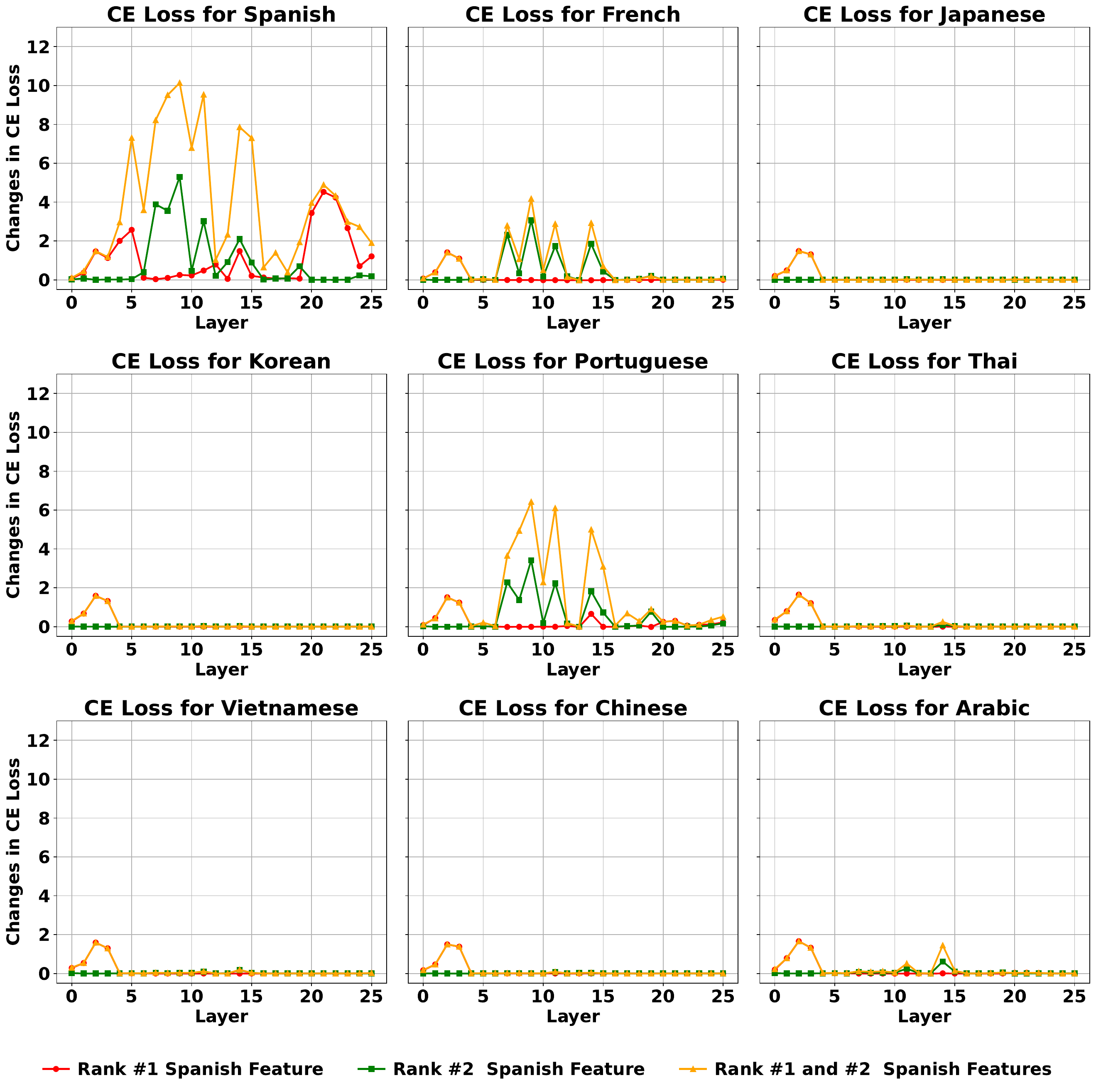}
  \caption{The change in CE loss for various languages after ablating Spanish features for Gemma 2 2B.}
  \label{fig:line_gemma2_0}
\end{figure*}

\begin{figure*}[t] 
\centering
  \includegraphics[width=\textwidth]{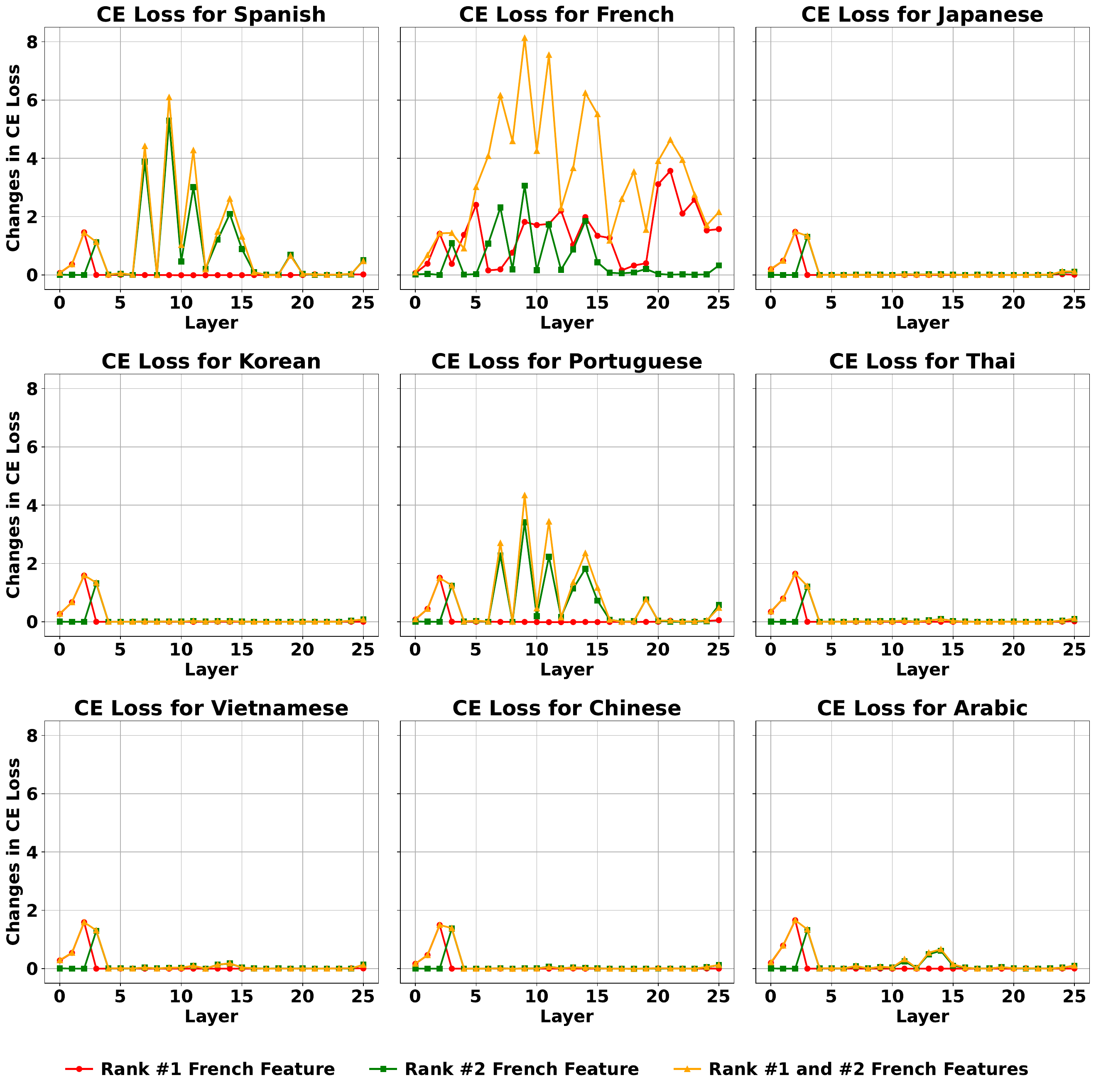}
  \caption{The change in CE loss for various languages after ablating French features for Gemma 2 2B.}
  \label{fig:line_gemma2_1}
\end{figure*}

\begin{figure*}[t] 
\centering
  \includegraphics[width=\textwidth]{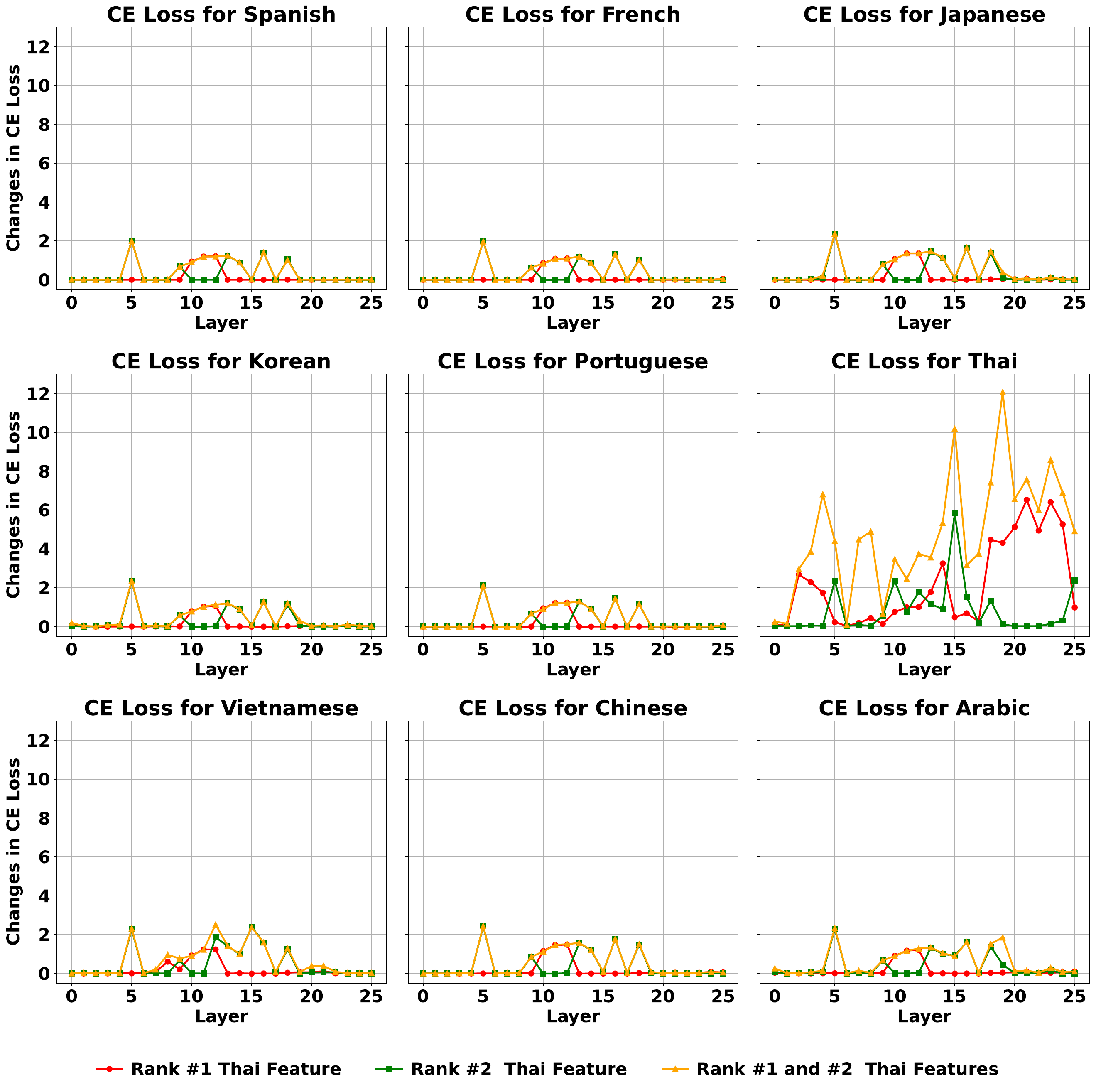}
  \caption{The change in CE loss for various languages after ablating Thai features for Gemma 2 2B.}
  \label{fig:line_gemma2_2}
\end{figure*}

\begin{figure*}[t] 
\centering
  \includegraphics[width=\textwidth]{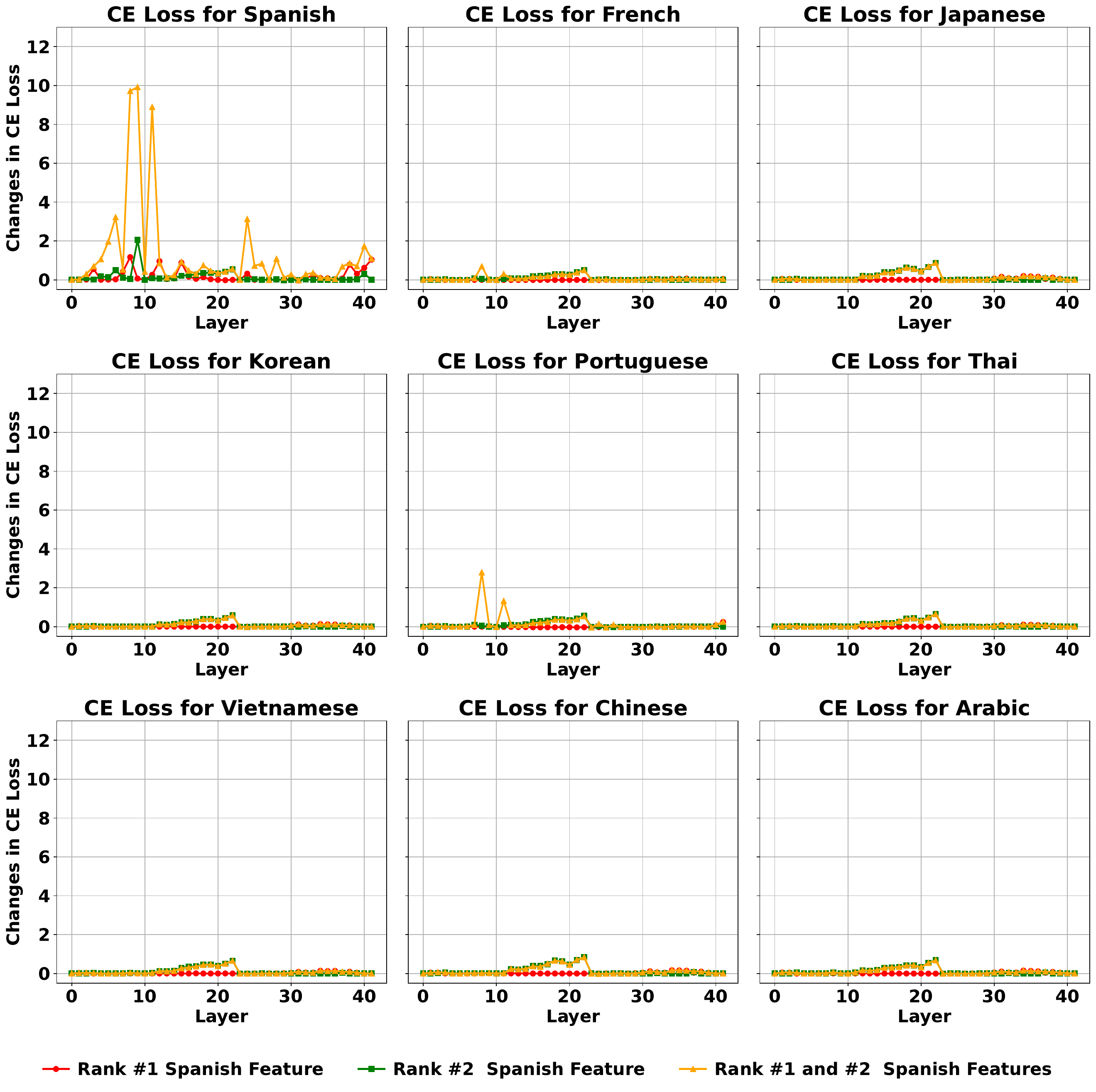}
  \caption{The change in CE loss for various languages after ablating Spanish features for Gemma 2 9B.}
  \label{fig:line_gemma9_0}
\end{figure*}

\begin{figure*}[t] 
\centering
  \includegraphics[width=\textwidth]{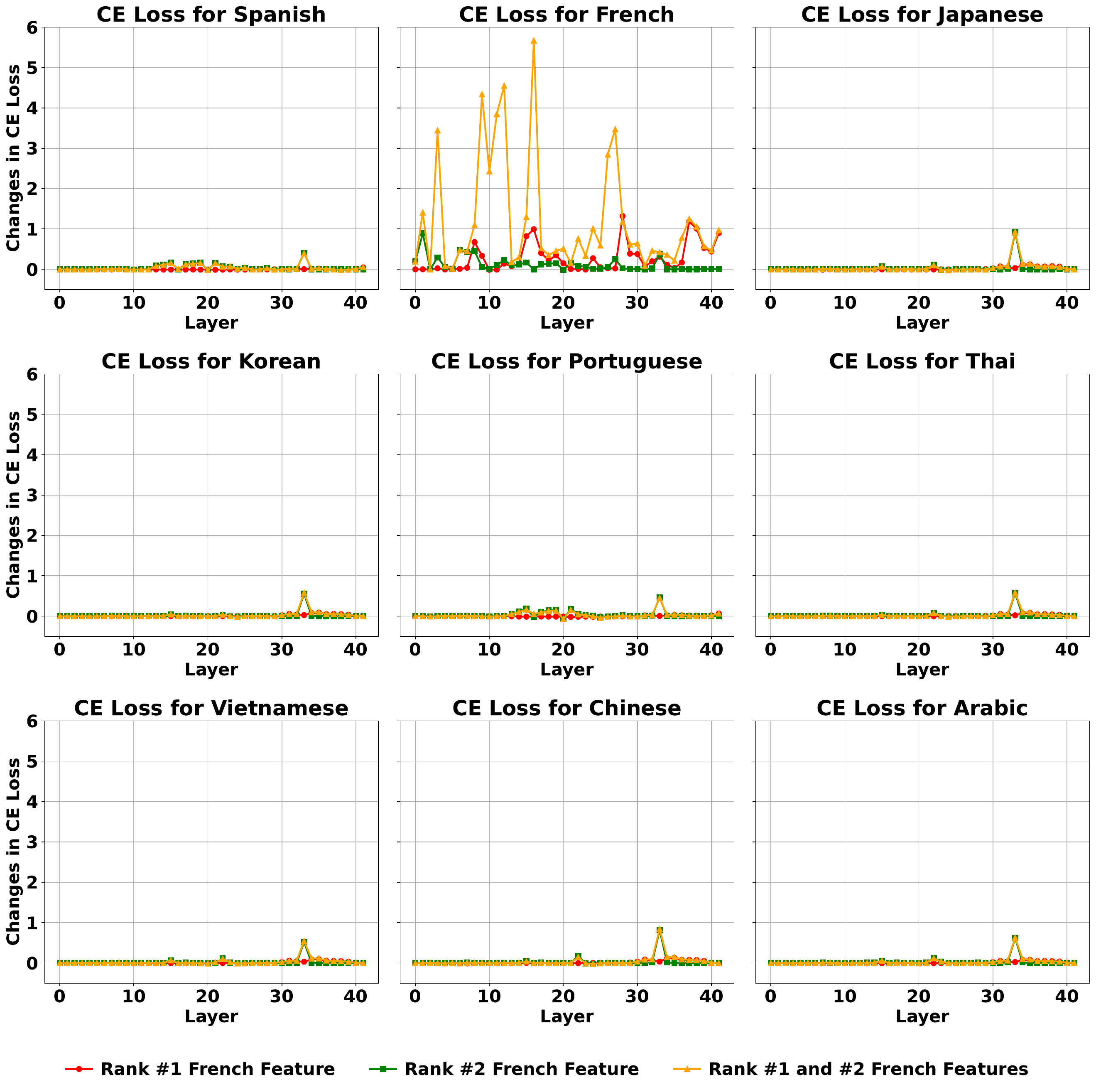}
  \caption{The change in CE loss for various languages after ablating French features for Gemma 2 9B.}
  \label{fig:line_gemma9_1}
\end{figure*}

\begin{figure*}[t] 
\centering
  \includegraphics[width=\textwidth]{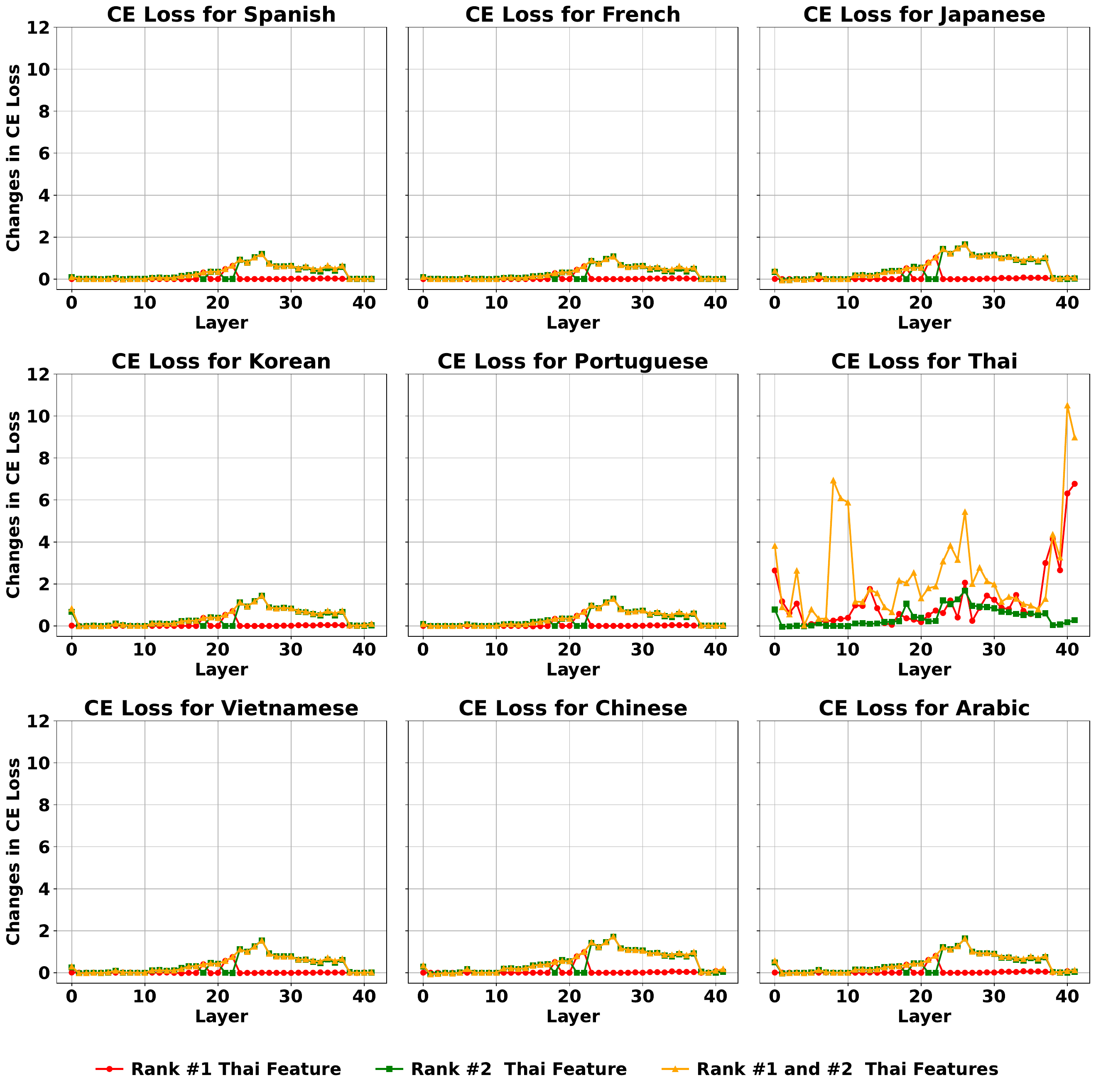}
  \caption{The change in CE loss for various languages after ablating Thai features for Gemma 2 9B.}
  \label{fig:line_gemma0_2}
\end{figure*}

\begin{figure*}[t] 
\centering
  \includegraphics[width=\textwidth]{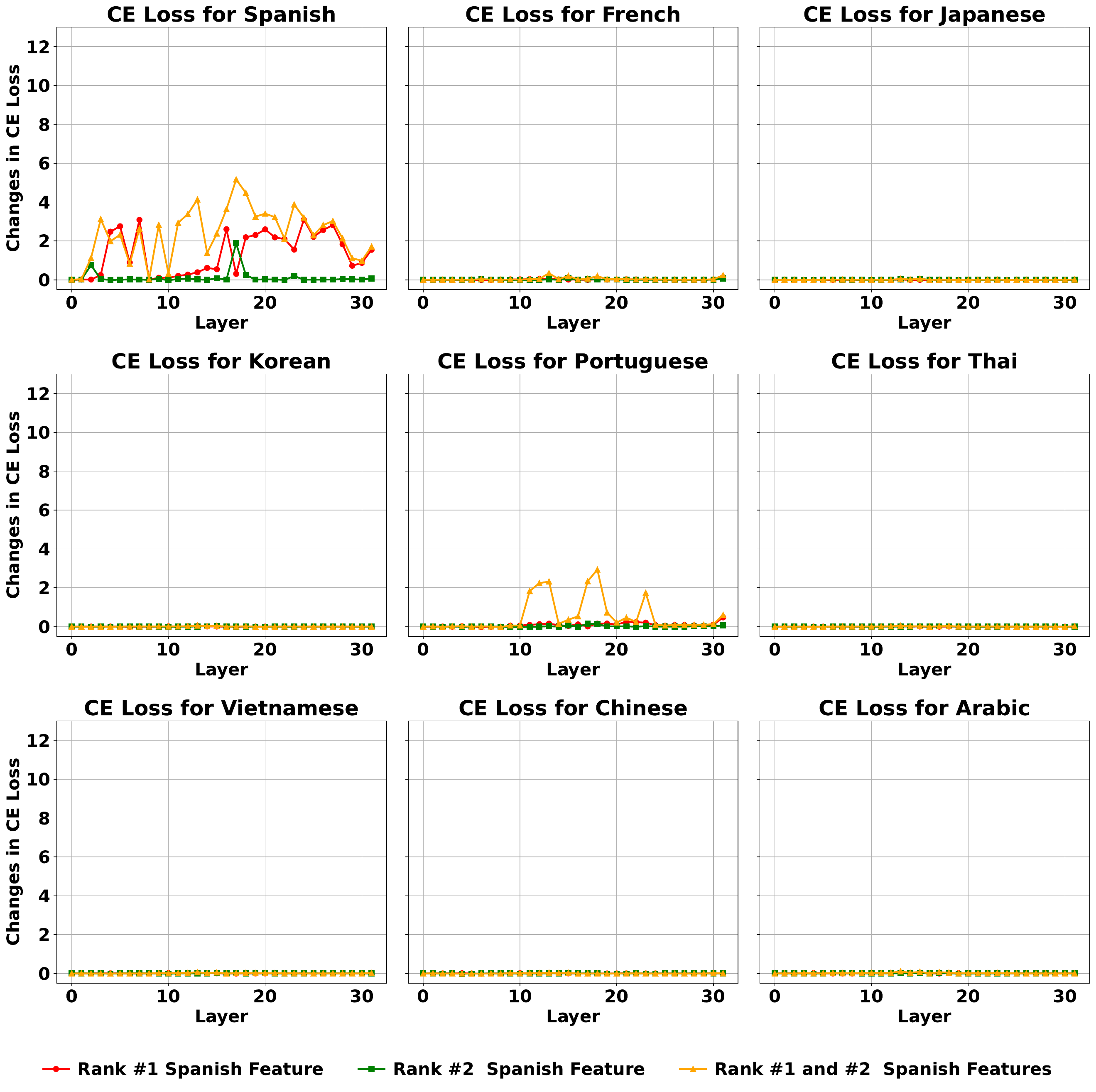}
  \caption{The change in CE loss for various languages after ablating Spanish features for Llama-3.1-8B.}
  \label{fig:line_Llama-0}
\end{figure*}

\begin{figure*}[t] 
\centering
  \includegraphics[width=\textwidth]{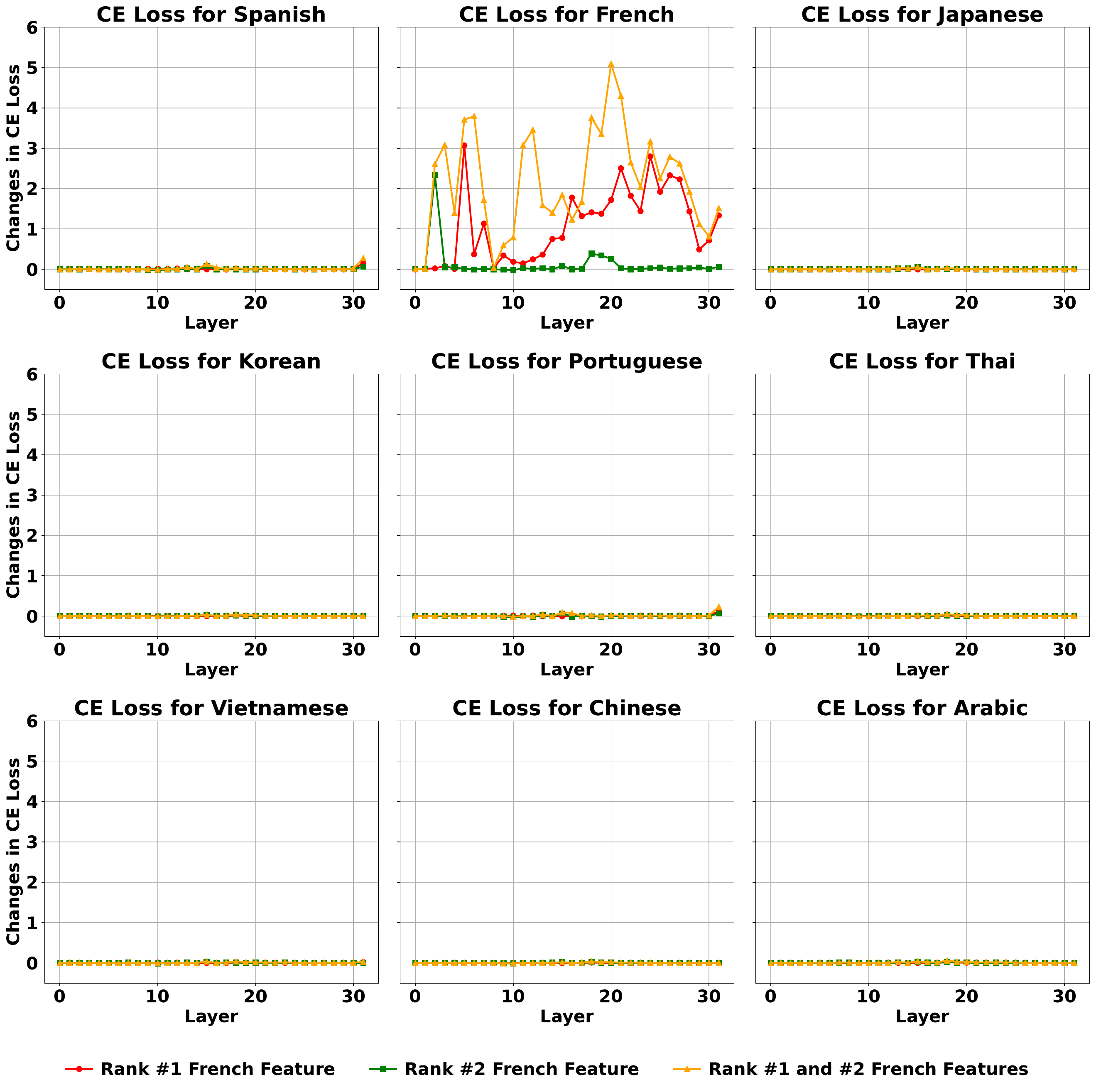}
  \caption{The change in CE loss for various languages after ablating French features for Llama-3.1-8B.}
  \label{fig:line_Llama-1}
\end{figure*}

\begin{figure*}[t] 
\centering
  \includegraphics[width=\textwidth]{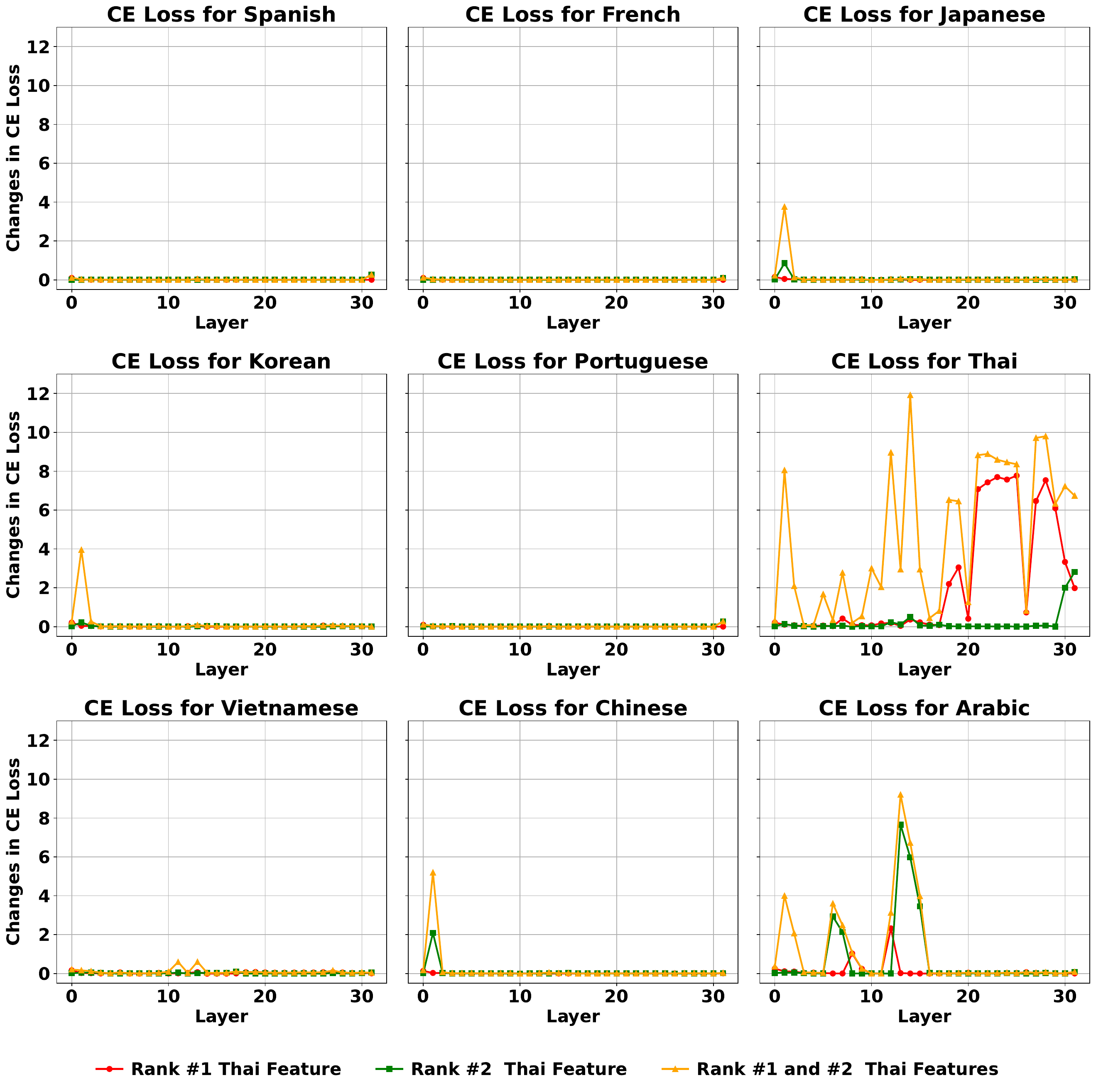}
  \caption{The change in CE loss for various languages after ablating Thai features for Llama-3.1-8B.}
  \label{fig:line_Llama-2}
\end{figure*}

Additional results for multiple features are demonstrated in Figure~\ref{fig:line_gemma2_0}-\ref{fig:line_Llama-2}. The results are similar to those in Figure~\ref{fig:2_top_features}.

\end{document}